%% file: PaperForReview.tex
\documentclass[10pt,twocolumn, letterpaper]{article}

\usepackage{cvpr}              % To produce the CAMERA-READY version
\usepackage{graphicx}
\usepackage{amsmath}
\usepackage{amssymb}
\usepackage{booktabs}
\usepackage[table]{xcolor}
\usepackage{makecell}
\usepackage{lipsum}
\usepackage{amsmath}
\usepackage[linesnumbered,ruled,vlined]{algorithm2e}

\SetKwComment{Comment}{/* }{ */}
\usepackage{dsfont}
\usepackage{makecell} % Include makecell package for advanced table cell formatting
\usepackage{multicol}
\usepackage{changepage} % For adjustwidth environment
\usepackage{pgfplots}
\pgfplotsset{compat=1.17}
\usepackage{booktabs} % For prettier tables
\usepackage{siunitx} % For aligning numbers by decimal points

\usepackage{graphicx}
\usepackage{geometry}
\usepackage{setspace}
\usepackage{tcolorbox}
\usepackage{amsmath}
\usepackage{array}
\usepackage{booktabs}
\usepackage{colortbl}
\usepackage{caption}
\usepackage{adjustbox}
\usepackage{makecell}
\usepackage{float}

\usepackage{amsmath}
\usepackage{array}
\usepackage{fixltx2e}
\usepackage{dblfloatfix}

\usepackage{enumitem}
\usepackage{xcolor}
\usepackage{tikz}
\usepackage{setspace}
\usepackage{tcolorbox}
\usepackage{amsmath}
\usepackage{array}
\usepackage{booktabs}
\usepackage{colortbl}
\usepackage{caption}
\usepackage{adjustbox}
\usepackage{makecell}
\usepackage{float}
\usepackage{multicol}
\usepackage{changepage} % For adjustwidth environment
\usetikzlibrary{patterns}
\usepackage{MnSymbol}
\usepackage{wasysym}
\usepackage{listings}
\usepackage{mdframed}

\usepackage{pgfplots}
\usetikzlibrary{arrows.meta}
\usepgfplotslibrary{groupplots}
\usepgfplotslibrary{dateplot}

\pgfplotsset{compat=newest,
    width=6cm,
    height=3cm,
    scale only axis=true,
    max space between ticks=25pt,
    try min ticks=5,
    every axis/.style={
        axis y line=left,
        axis x line=bottom,
        axis line style={thick,->,>=latex, shorten >=-.4cm}
    },
    every axis plot/.append style={thick},
    tick style={black, thick}
}
\tikzset{
    semithick/.style={line width=0.8pt},
}

\include{utils/defs}

\usepackage[pagebackref,breaklinks,colorlinks]
{hyperref}

\usepackage[capitalize]{cleveref}
\crefname{section}{Sec.}{Secs.}
\Crefname{section}{Section}{Sections}
\Crefname{table}{Table}{Tables}
\crefname{table}{Tab.}{Tabs.}

\def\cvprPaperID{*****} % *** Enter the CVPR Paper ID here
\def\confName{CVPR}
\def\confYear{2022}

\begin{document}

%%%%%%%%% TITLE - PLEASE UPDATE
\title{Ensemble Modeling of Multiple Physical Indicators to Dynamically Phenotype Autism Spectrum Disorder}

% \author{Marie Huynh\\
% Stanford University\\
% Institution1 address\\
% {\tt\small mahuynh@stanford.edu}
% % For a paper whose authors are all at the same institution,
% % omit the following lines up until the closing ``}''.
% % Additional authors and addresses can be added with ``\and'',
% % just like the second author.
% % To save space, use either the email address or home page, not both
% }
\newcommand*\samethanks[1][\value{footnote}]{\footnotemark[#1]}
\author{Marie Huynh 
\thanks{\{mahuynh,akline,mourya,kaiti.dunlap,
cezmi,mhonar,azizian,dpwall\}@stanford.edu, 
Stanford University} \and Aaron Kline \samethanks \and Saimourya Surabhi \samethanks \and Kaitlyn Dunlap \samethanks \and Onur Cezmi Mutlu \samethanks \and Mohammadmahdi Honarmand \samethanks \and Parnian Azizian \samethanks \and Peter Washington \thanks{pyw@hawaii.edu, University of Hawaii}\and Dennis P. Wall\samethanks[1] }

\maketitle
%%%%%%%%% ABSTRACT

\begin{abstract}
Early detection of autism, a neurodevelopmental disorder marked by social communication challenges, is crucial for timely intervention. Recent advancements have utilized naturalistic home videos captured via the mobile application GuessWhat. Through interactive games played between children and their guardians, GuessWhat has amassed over 3,000 structured videos from 382 children, both diagnosed with and without Autism Spectrum Disorder (ASD). This collection provides a robust dataset for training computer vision models to detect ASD-related phenotypic markers, including variations in emotional expression, eye contact, and head movements. We have developed a protocol to curate high-quality videos from this dataset, forming a comprehensive training set. Utilizing this set, we trained individual LSTM-based models using eye gaze, head positions, and facial landmarks as input features, achieving test AUCs of 86\%, 67\%, and 78\%, respectively. To boost diagnostic accuracy, we applied late fusion techniques to create ensemble models, improving the overall AUC to 90\%. This approach also yielded more equitable results across different genders and age groups. Our methodology offers a significant step forward in the early detection of ASD by potentially reducing the reliance on subjective assessments and making early identification more accessibly and equitable.

\end{abstract}

%%%%%%%%% BODY TEXT
\section{Introduction}
%\subsection{Background}
\label{sec:intro}

Autism Spectrum Disorder (ASD) is a complex neurodevelopmental condition that affects approximately 1 in 36 children across diverse ethnic, racial, and socioeconomic backgrounds, highlighting its widespread impact and the need for comprehensive strategies to address it \cite{maenner2023prevalence}. Children with ASD face significant challenges in communication, social interactions, repetitive behaviors, and restricted interests, often leading to profound difficulties in everyday functioning and development \cite{johnson2007identification}. The impact of ASD extends beyond individual health, imposing a substantial economic burden with an estimated lifetime social cost of approximately \$3.6 million per affected individual \cite{cakir2020lifetime}. 

Early interventions are crucial as the developmental gap between children with ASD and neurotypical (NT) children widens over time \cite{dawson2003early}. Early diagnosis can lead to improved health outcomes; however, despite the potential for reliable diagnosis as early as 16–24 months, the average age of diagnosis is 4.5 years \cite{lord2006autism, shen2020biomarkers}. Diagnostic processes typically involve long waitlists and assessments, resulting in an average delay of two years \cite{gordon2016whittling}. Current diagnoses rely on in-person behavioral assessments, which are costly and subjective, lacking definitive medical tests or biomarkers \cite{first2013diagnostic}. This subjective process introduces variability and the potential for misdiagnosis influenced by clinician experience, training, and social biases \cite{mazefsky2006discriminative}. 

Digital phenotyping using naturalistic home videos offers a promising approach for faster and more objective diagnosis of ASD and other developmental conditions. 

\section{Related Work}

\subsection{Digital Phenotyping of Autism}
Computer vision tools have shown promise in identifying multiple phenotypes such as emotion, eye movement, and posture tracking in children through video analysis \cite{Sapiro2019, lakkapragada2022classification, zhao2021identifying}. However, these tools often lack structured data on children with autism, and predictive models using individual phenotypes have rarely been assessed for fairness or integrated into ensemble models.

Eye tracking has been studied extensively for diagnosing ASD in children, with recent machine learning algorithms analyzing eye gaze patterns to differentiate between ASD and neurotypical (NT) children \cite{guclu2021machine, liaqat2021predicting, varma2021identification}.  In a meta-analysis of eye-tracking based ML models for distinguishing between ASD and NT individuals, Wei et al. \cite{guclu2021machine}reported a pooled classification accuracy of 81\%, specificity of 79\%, and sensitivity of 84\%. However, Takahashi and Hashiya \cite{takahashi2019advanced} reported high degress of  variation in eye-tracking data collection, emphasizing the need for additional data regularization methods as well as incorporation of other data modalities.

 Other modalities such as head movement \cite{zhao2022identifying}, facial expression \cite{jiang2019classifying, banerjee2023training},  finger movement \cite{simeoli2021using}, and facial engagement \cite{varma2021identification}  have shown valuable for identifying autism. Nonetheless, challenges related to data variability, generalizability across diverse populations, and integration with other diagnostic measures remain.

Building upon this groundwork, we use deep time-series models such as LSTM and GRU to analyze eye gaze, face, and head features to predict ASD. Our approach will include an advanced feature engineering pipeline and explore fusion methods to enhance predictive power and model robustness.

\subsection{Data Fusion For ASD Classification}

Recent studies have focused on the potential of data fusion for improving ASD classification performance. Perochon et al. \cite{perochon2023early} recently evaluated an autism screening digital application administered during pediatric well-child visits for children aged 17–36 months. Their algorithm integrated multiple digital phenotypes, achieving an AUROC of 0.90, sensitivity of 87.8\%, specificity of 80.8\%, negative predictive value of 97.8\%, and positive predictive value of 40.6\%. These findings underscore the potential of combining data sources to improve diagnostic outcomes.

Further extending this line of work, Thompson et al. \cite{thompson2024integration} investigated the integration of acoustic and linguistic markers with visual data, reporting enhancements in the system's ability to predict ASD traits in varied social contexts. The integration of physiological data has been further explored by Nakamura et al. \cite{nakamura2024physiological}, who demonstrated that combining heart rate variability and skin conductance with traditional behavioral assessments could offer a more comprehensive understanding of ASD.

Our research builds on these foundations, focusing on mobile videos captured in a home environment during gameplay. This approach offers a scalable and open method for dynamic data collection, crucial for conditions like autism that are not static \cite{washington2020data}. It enables continuous monitoring and authentic behavioral data, potentially facilitating earlier and more accurate diagnoses and personalized interventions. Moreover, our work emphasizes fusing multiple indicators—such as eye gaze, head positions, and facial landmarks—extracted from these videos to construct robust diagnostic models. By analyzing video data in real-world settings, we aim to bridge the gap between clinical practice and everyday environments, offering a practical solution for widespread autism screening and diagnosis.

\section{Dataset}

\label{sec:formatting}

\subsection{Data Source and Type}

As a home-based digital therapeutic for children with developmental 
delays, the mobile application GuessWhat 
\cite{kalantarian2019guess}
provides a rich dataset of videos to train new models to 
phenotype autism digitally. To date, GuessWhat has amassed more than 
3,000 highly structured videos of 382 children aged from 2 to 
12 years old with and without an autism diagnosis.

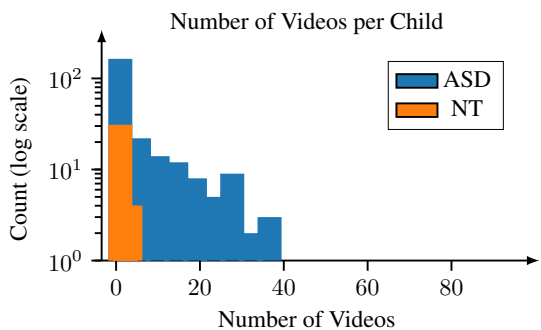
\begin{figure}[h]
    \centering
     \scalebox{.9}{\input{latex/figures/superusers_plot}}
    \caption{Presence of Superusers in the ASD Class. Some children with ASD dominate the data with dozens of videos.}
    \label{fig:superusers}
\end{figure}

%-----------------------------------------------------------------------

\subsection{Filtering Pipeline}

\begin{figure*}
    \centering
    \resizebox{0.5\textwidth}{!}{\input{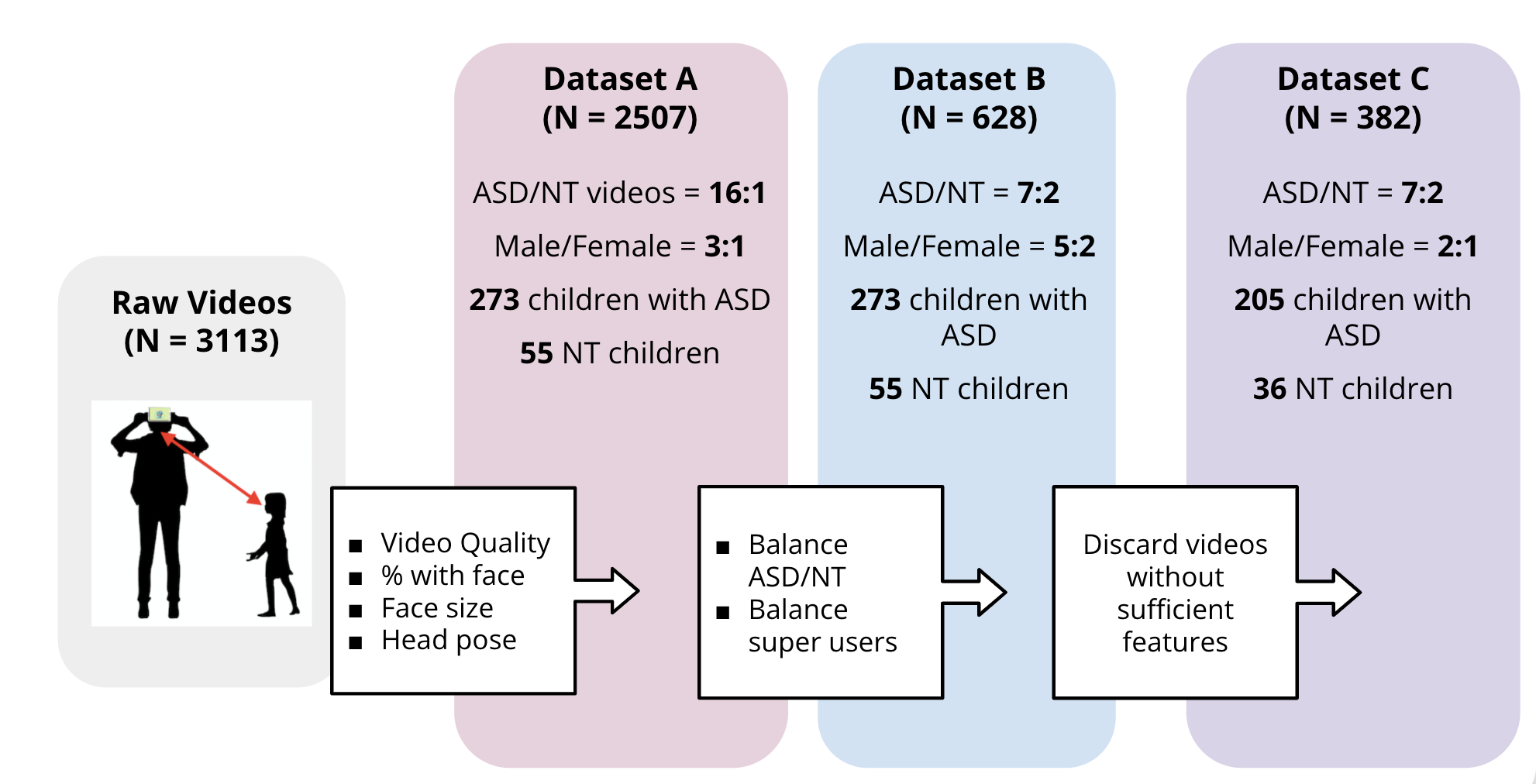}}
    \caption{Key Filtering Steps}
   \label{fig:filtering_steps}
\end{figure*}

The instability of  home videos can create noise and data drifts that reduce the performance of the models. Furthermore, some videos may be feature-poor, or the child of interest may be too far away to extract features of interest, etc. Rigorous filtering and feature engineering are needed to account for these limitations and build a minimally viable training set for our task. Our filtering pipeline can be summarized in three steps as illustrated in Figure~\ref{fig:filtering_steps}.

% GuessWhat mobile application has amassed around 4,348 raw videos to date. To produce information-rich feature generation sequences we used AWS. The criteria include videos of good quality for feature extraction, and each video to focus on the child of interest. After filtering, we narrowed the dataset down to 3,113 videos, from which we extract eye-gazing, face, and head features coordinates for each frame per video. 

% To ensure these two criteria, we filtered our videos by using the Amazon Rekognition Video API for face detection \cite{awsrekognition2023}, which returns estimates for sharpness, brightness, head pose, and face size values for each video. To meet the first criterion, we filtered videos to guarantee clarity, using sharpness and brightness metrics, featured a sufficiently large face for detecting eye-gazing features, had the eyes open for more than 70\% of the video duration, and maintained the face predominantly facing the camera (as indicated by the head pose). To ensure the second criterion, we selected videos where the proportion of the face was large enough to facilitate reliable ASD predictions and where the presence of multiple faces was minimal, ensuring the focus remained on a single child. This filtered dataset, which we will denote Dataset A, counts 2123 videos, including 288 children, as shown in Table~\ref{tab::datasetA}. 

The criteria for the videos are as follows: (1) they must be of high quality, suitable for feature extraction, and (2) each video must focus on the child of interest. To ensure these criteria, we filtered our videos using the Amazon Rekognition Video API for face detection \cite{awsrekognition2023}, which provides estimates for sharpness, brightness, head pose, facial landmarks, and face size for each video. For the first criterion, we selected videos that guaranteed clarity through sharpness and brightness metrics, featured a sufficiently large face for detecting eye-gazing features, had the eyes open for more than 70\% of the video's duration, and maintained the face predominantly facing the camera (as indicated by the head pose). To meet the second criterion, we selected videos where the face was proportionally large enough for reliable feature extraction and where the presence of multiple faces was minimal, ensuring the focus remained on a single child. This filtered dataset, referred to as Dataset A, includes 2123 videos featuring 288 children, as shown in Table~\ref{tab::datasetA}.

% It is important to notice the significant imbalance between the ASD and NT classes and the presence of super-users in the ASD class.\\ 

\begin{table}[h]
\centering
\begin{tabular}{lS[table-format=4.0]S[table-format=2.1]}
\toprule
{} & {Number of videos} & {Number of Children} \\
\midrule
ASD & 2007 & 245\\
NT      & 116                 &    43       \\
\textbf{Total}      & 2123              & 288        \\
\bottomrule
\end{tabular}
\caption{Distribution of ASD Labels - Dataset A}
\label{tab::datasetA}
\end{table}

\subsection{Bias and Imbalances}

Our dataset contains a significant imbalance with respect to ASD and NT classes, as reflected in Table ~\ref{tab::datasetA}. Dataset A has an ASD to NT ratio of 17:1. Furthermore, some users in the ASD class are more represented than others due to more gameplay, as can be seen in Figure~\ref{fig:superusers}. 

Since we only have 116 videos for NT children (cf. Table ~\ref{tab::datasetA}), we manually inspected all the NT videos to filter out those of poor quality or that did not adhere to the specified constraints. During this review, 7 videos were identified as invalid for reasons such as the parent playing and the child holding the phone, siblings playing one after the other in the same video, etc.

To avoid any bias towards a particular child and mitigate the imbalance between ASD and NT, we under-sampled the ASD class by keeping at most two videos with the highest quality (mean of sharpness and brightness) for each child. We kept all of our data for NT children (since it is a minority class). We obtained dataset B, which contains 700 videos (cf. Figure~\ref{fig:filtering_steps}).

\subsection{Feature Extraction}
The input of our models consists of a sequence of k frames, where each frame contains a feature vector of size d (d being the dimension of features for a given modality, $d \in \{2, 7, 60\}$ for the eye gazing, head and face modality, respectively). Hence, our input is a multivariate time-series denoted as $[X^1, ..., X^k]$ where $X^i = [x^i_1, ..., x^i_d]$ and our output is a binary label Y, where $Y \in \{0, 1\}$ (0: NT, 1: ASD).  

\begin{figure*}
    \centering
    \resizebox{0.5\textwidth}{!}{\input{latex/figures/feature_extraction}}
    \caption{Feature Extraction Scheme. Every feature vector comes with a confidence score ranging from 0 to 100.}
    \label{fig:extraction}
\end{figure*}
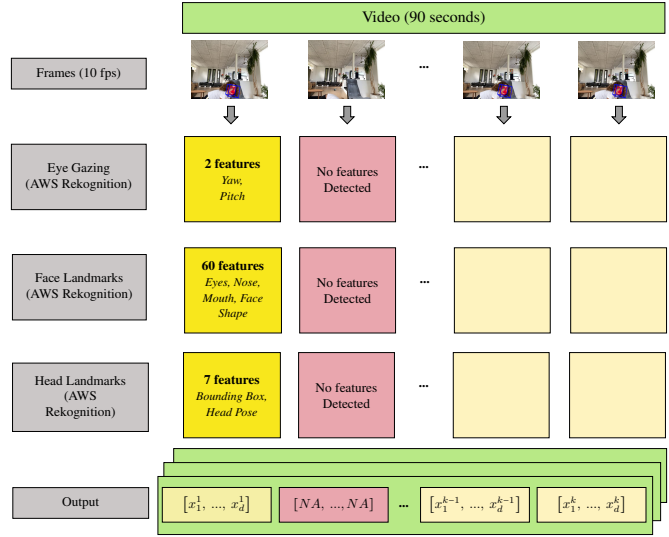

GuessWhat videos have an average frame rate of 28 frames per second and average length of 90 seconds. We down-sampled the dataset B videos to 10 frames per second and utilized AWS Rekognition\cite{awsrekognition2023} to extract frame-level eye gaze, head pose, and facial features, as illustrated in Figure~\ref{fig:extraction}. Rekognition-provided confidence scores demonstrated a mean confidence of 79.4 for eye and 99.12 for face and head detection features for every frame with a face. 
After extraction, we dropped any video with less than 15 seconds of features extracted and obtained a high-quality Dataset C with 688 videos (cf. Figure~\ref{fig:filtering_steps}). Per-video and per-child demographics of the final dataset are detailed in Table ~\ref{tab:demographics}.

\begin{table}[ht]
    \centering
    \resizebox{8cm}{!}{
    \begin{tabular}{lcccccc}
        \toprule
        \textbf{Demographic} & \multicolumn{3}{c}{\textbf{Video Level}} & \multicolumn{3}{c}{\textbf{Child Level}} \\
        \cmidrule(lr){2-4} \cmidrule(lr){5-7}
        & \textbf{ASD} & \textbf{NT} & \textbf{Total} & \textbf{ASD} & \textbf{NT} & \textbf{Total}  \\
        \midrule
        \multicolumn{7}{c}{\textbf{Gender}} \\
        \midrule
        Male               & 426 & 53 & 479  & 178 & 21 & 199  \\
        Female             & 124 & 55 & 179  & 53 & 21 & 74  \\
        NA                 & 30  & 0 & 30   & 12 & 0 & 12  \\
        \midrule
        \multicolumn{7}{c}{\textbf{Age}} \\
        \midrule
        1-4               & 251 & 51 & 302  & 101 & 18 & 119  \\
        5-8               & 194 & 33 & 227  & 84  & 14 & 98  \\
        9-12              & 135 & 24 & 159  & 58  & 10 & 68 \\
        \midrule
        \multicolumn{7}{c}{\textbf{Location}} \\
        \midrule
        Unknown & 258 & 90 &  348 &   109 &  32 & 141\\
        United States & 277 & 4 &  281 & 114 & 3 &  117 \\
        Outside US & 45 & 14 & 59  & 20 &   7 & 27\\
        \bottomrule
    \end{tabular}
    }
    \caption{Demographics Statistics at \textbf{the Video Level} and \textbf{the Child Level}}
    \label{tab:demographics}
    
\end{table}
\vspace{-10pt}
%----------------------------------------------------------------------
\section{Experimental Setup}

\subsection{Data Preprocessing}
\begin{figure*}[h]
    \centering
    \hspace{-2cm} 
    \resizebox{0.6\textwidth}{!}{\input{latex/figures/feature_engineering}}
    \caption{Key Feature Engineering Steps}
    \label{fig:feature_engineering} % Ensure label is unique
\end{figure*}
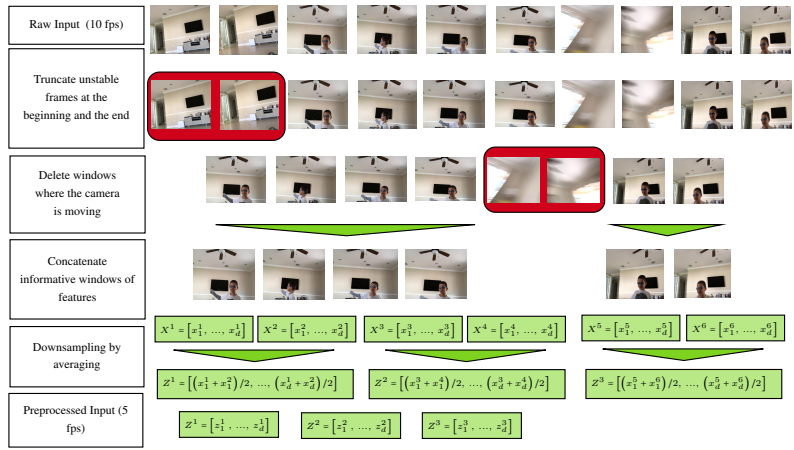

Mobile data is inherently noisy, requiring robust feature engineering to extract informative sequences for learning. Our  feature engineering pipeline, depicted in Figure~\ref{fig:feature_engineering}, addresses several challenges.

At the start and end of videos, our features of interest are often missing due to camera stabilization or game initiation with the child. To mitigate this, we truncate frames where no face is detected.

\begin{figure}[h]
    \centering
    \includegraphics[width=\linewidth]{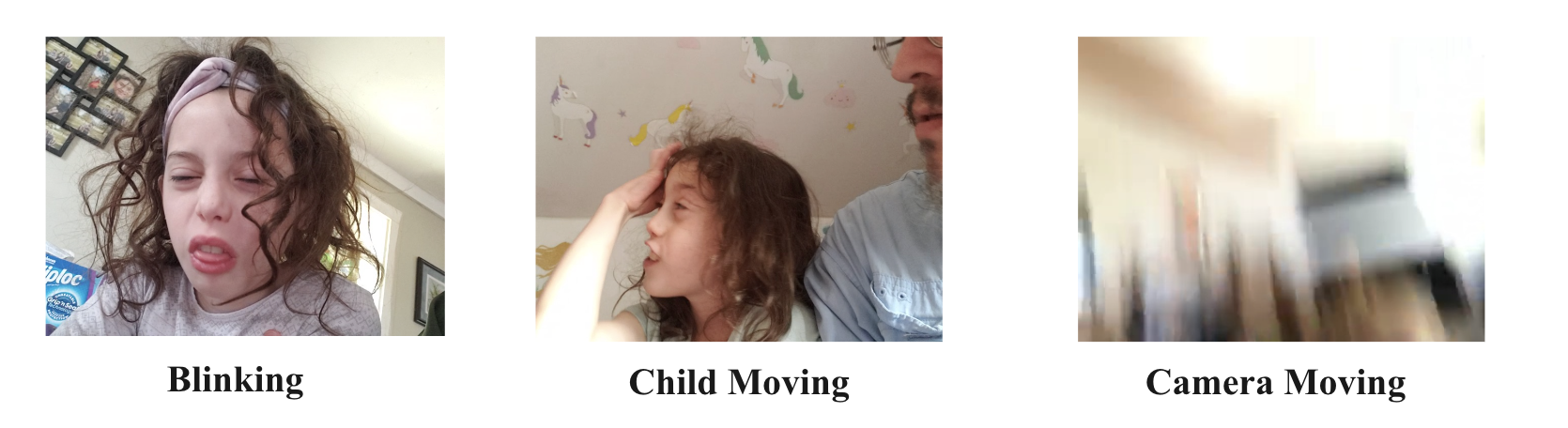}
    \caption{Discontinuities Factors}
    \label{fig:missingness} % Ensure label is unique
\end{figure}

Within videos, we identify periods of missing data, illustrated in Figure~\ref{fig:missingness}. 
\begin{figure*}
    \centering
    \resizebox{0.5\textwidth}{!}{\input{normalization}}
    \caption{Normalization and Missingness Examples for Head Features.}
    \label{fig:norm_missing}
\end{figure*}

Missing eye gaze data occurs when the child's eyes are closed, not facing the camera, or when the camera angle is off. These instances can be informative, reflecting the child's interaction challenges or meaningful movements. Conversely, periods without face detection or poorly centered cameras provide no useful information. We exclude these uninformative windows and concatenate informative segments to enhance feature continuity.

To reduce input length,  we averaged features every two frames, for an effective frame rate of 5 fps, as shown in Figure~\ref{fig:feature_engineering}. We then normalized the features and represented missing frames as a vector of tokens (-1) as shown in Figure \ref{fig:norm_missing}., in order to incorporate missingness as a feature in the temporal data structure. 

\vspace{-10pt}

\subsection{Model Training}
The resulting dataset was split at the child level into training, validation, and test sets to prevent data leakage, especially for children with multiple videos. Split details are summarized in Table \ref{tab:splits}. To ensure fairness and representativeness, we stratified the splits by age group (1-4, 5-8, 9-12) and gender (Male, Female, Other), summarized in  Table \ref{tab:demographics_splits}. This approach may result in an unconventional data distribution but ensures consistent and accurate representation across these demographic factors.

% \begin{table}[h]
%   \centering
%   \scalebox{1.1}{
%     \begin{tabular}{|c|c|c|c|c|}
%     \hline
%     \cellcolor{black!55} & \multicolumn{2}{c|}{\makecell{Number of\\Videos}} & \multicolumn{2}{c|}{\makecell{Number of\\Children}} \\
%     \hline
%     \cellcolor{black!55} & \textbf{ASD} & \textbf{NT} & \textbf{ASD} & \textbf{NT} \\
%     \hline
%     Train (63.4\%) & 240 & 46 & 114 & 18 \\
%     Train Upsampled & 240 &70 & 114 & 18 \\
%     Val (15.2\%) & 68 & 17 & 36 & 8 \\
%     Test (21.4\%) & 75 & 17 & 37 & 8 \\
%     \hline
%     \end{tabular}
%   }
%   \caption{Statistics of Train/Val/Test Splits. These videos have the three modalities (eye, face, head). We upsample the NT class in Train Upsampled.}
% \end{table}

\begin{table}[h]
  \centering
  \resizebox{5cm}{!}{
    \begin{tabular}{|c|c|c|c|c|}
    \hline
    \cellcolor{black!55} & \multicolumn{2}{c|}{\makecell{Number of\\Videos}} & \multicolumn{2}{c|}{\makecell{Number of\\Children}} \\
    \hline
    \cellcolor{black!55} & \textbf{ASD} & \textbf{NT} & \textbf{ASD} & \textbf{NT} \\
    \hline
    Train (62.2\%) & 363 & 65 & 148 & 23 \\
    Train Upsampled & 363 & 148 & 97 & 23 \\
    Val (18\%) & 105 & 19 & 49 & 8 \\
    Test (19.8\%) & 112 & 24 & 46 & 11 \\
    \hline
    \end{tabular}
  }
  \caption{Statistics of Train/Val/Test Splits. These videos have the three modalities (eye, face, head).}
  \label{tab:splits}
\end{table}
\vspace{-5pt}
% \begin{table}[ht]
%     \centering
%     \begin{tabular}{l c c c}
%         \toprule
%         \textbf{Demographic} & \textbf{TRAIN} & \textbf{TEST} & \textbf{VAL} \\
%         \midrule
%         \multicolumn{4}{c}{\textbf{Age}} \\
%         \midrule
%         1-4 & 130 & 39 & 37 \\
%         5-8 & 102 & 26 & 29 \\
%         9-12 & 54 & 28 & 18 \\
%         \midrule
%         \multicolumn{4}{c}{\textbf{Gender}} \\
%         \midrule
%         Male & 188 & 65 & 50 \\
%         Female & 87 & 25 & 28 \\
%         \midrule
%         \multicolumn{4}{c}{\textbf{Location}} \\
%         \midrule
%         Unknown & 155 & 36 & 52 \\
%         United States & 109 & 47 & 26 \\
%         Outside US & 22 & 10 & 6 \\
%         \bottomrule
%     \end{tabular}
%     \caption{Demographics of the Splits}
%     \label{tab:demographics_splits}
% \end{table}

\begin{table}[ht]
    \centering
    \resizebox{5cm}{!}{
    \begin{tabular}{l c c c}
        \toprule
        \textbf{Demographic} & \textbf{TRAIN} & \textbf{TEST} & \textbf{VAL} \\
        \midrule
        \multicolumn{4}{c}{\textbf{Age}} \\
        \midrule
        1-4 & 198 & 58 & 46 \\
        5-8 & 137 & 44 & 46 \\
        9-12 & 93 & 34 & 32 \\
        \midrule
        \multicolumn{4}{c}{\textbf{Gender}} \\
        \midrule
        Male & 286 & 99 & 94 \\
        Female & 119 & 34 & 26 \\
        None/Other & 23 & 3 & 4 \\
        \midrule
        \multicolumn{4}{c}{\textbf{Location}} \\
        \midrule
        Unknown & 225 & 68 & 55 \\
        United States & 166 & 60 & 55 \\
        Outside US & 37 & 8 & 14 \\
        \bottomrule
    \end{tabular}}
    \caption{Demographics of the Splits}
    \label{tab:demographics_splits}
\end{table}
\vspace{-10pt}

We trained LSTM and GRU models for the binary prediction tasks using eye gazing, face, and head features, all implemented in PyTorch. We hypothesized that fusing all three modalities (eye, head, face) would enhance predictive power, we tested two ensemble models: late fusion and intermediate fusion\cite{NEURIPS2019_9015}. For late fusion, we averaged the scores and used a linear layer,  concatenating logit outputs followed by a linear layer, to predict ASD probability. The intermediate fusion model concatenates the final hidden dimensions of the pre-trained models and processes them through a multi-layer perceptron, with both models using binary cross-entropy with class weights as their loss function. 
Given the dataset class imbalance, we also explored utilizing focal loss\cite{lin2017focal} but found performance similar to unmodified binary cross-entropy. We implemented early stopping with the default parameters of 3 epochs and delta of 0.001.
%\subsection{Models Hyperparameters}
%The optimal hyperparameters identified through optimization using the Optuna framework \cite{optuna_2019} are detailed in Table ~\ref{tab:hyperparameters}. Each modality's best model was selected based on the highest validation F1-score among 40 trials conducted. Tables \ref{tab:intermediate_fusion} and \ref{tab:late_fusion_linear} summarize the hyperparameters for the late fusion (Linear) and intermediate fusion models.

\section{Results and Discussion}
\subsection{Effect of Feature Engineering}
Table \ref{tab:feature_engineering_effect} shows the impact of feature engineering on the performance of three models (Eye, Head, Face) by comparing their AUC and F1-scores before and after applying the pipeline. The Eye model shows the most significant improvementthough its F1-score slightly decreases, indicating some trade-offs. The Face model exhibits minimal improvement, with its AUC modestly rising and the F1-score decreasing , showing limited effectiveness of the feature engineering.

\begin{table}[h]
\centering
\resizebox{5cm}{!}{
\begin{tabular}{|l|c|c|}
\hline
\textbf{Model} & \textbf{AUC Score} & \textbf{F1-Score (MA)} \\
\hline
Eye (Raw) & 0.66 & 0.66 \\
Eye (After) & \textbf{0.86} & \textbf{0.73} \\
\hline
Head (Raw) & 0.66 & 0.66 \\
Head (After) & \textbf{0.78} & \textbf{0.63} \\
\hline
Face (Raw) & 0.63 &  0.69\\
Face (After) & \textbf{ 0.67} & \textbf{0.63} \\
\hline
\end{tabular}}
\caption{Effect of Feature Engineering Pipeline on Model Performance. Each model is the best model tuned out of 40 trials, chosen on val F1-score.}
\label{tab:feature_engineering_effect}
\end{table}

\subsection{Performance comparison of our models}

\begin{figure*}[h]  % Using figure* to span two columns in a two-column document layout
    \centering
    % First row
    \begin{subfigure}[b]{0.4\textwidth}
        \includegraphics[width=\textwidth]{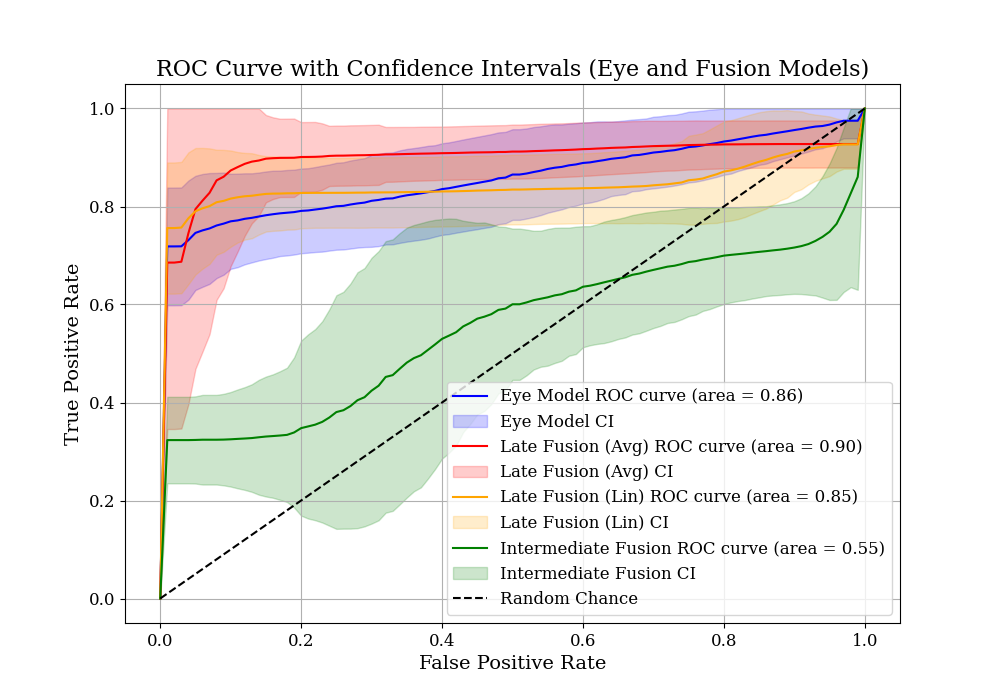}
        \caption{Receiver Operation Characteristic Curves (Eye and Fusion Models).}
        \label{fig:roc_curves_fusion}
    \end{subfigure}
    \hfill
    \begin{subfigure}[b]{0.4\textwidth}
        \includegraphics[width=\textwidth]{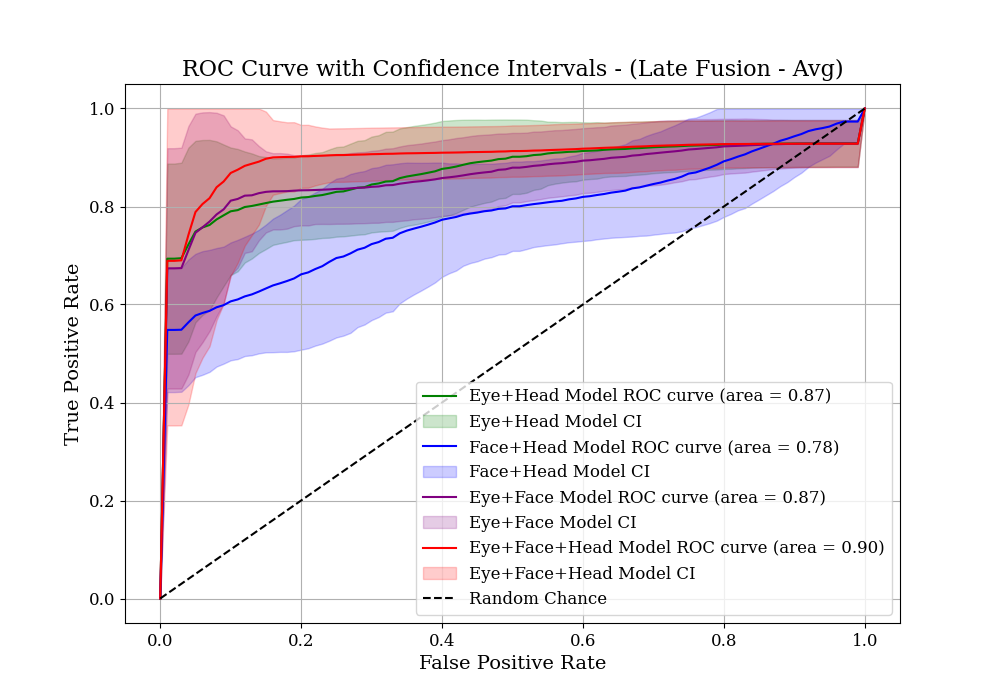}
        \caption{Receiver Operation Characteristic Curves - Combinations (Average).}
        \label{fig:roc_curves_avg_combinations}
    \end{subfigure}

    % Second row
    \bigskip  % Adds vertical space between the rows of subfigures
    \begin{subfigure}[b]{0.4\textwidth}
        \includegraphics[width=\textwidth]{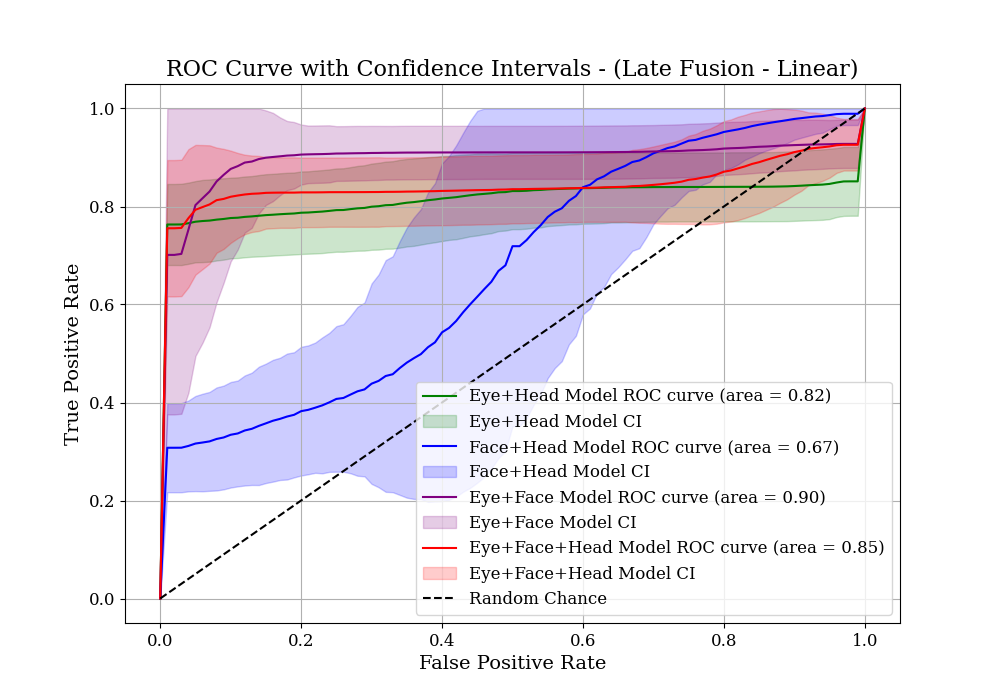}
        \caption{Receiver Operation Characteristic Curves - Combinations (Linear).}
        \label{fig:roc_curves_linear_combinations}
    \end{subfigure}
    \hfill
    \begin{subfigure}[b]{0.4\textwidth}
        \includegraphics[width=\textwidth]{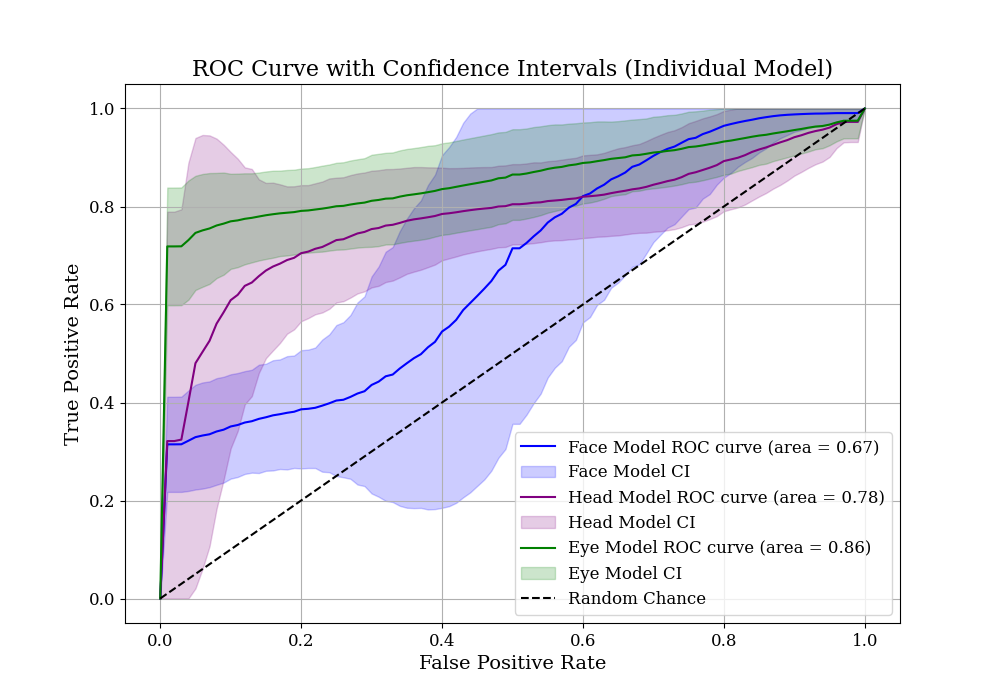}
        \caption{Receiver Operation Characteristic Curves (Individual Models).}
        \label{fig:roc_curves_individuals}
    \end{subfigure}

    \caption{Comprehensive ROC Curves Analysis}
    \label{fig:comprehensive_roc}
\end{figure*}

The eye and head models achieved strong predictive power with test AUCs of 0.86 and 0.78, respectively (Figure \ref{fig:roc_curves_individuals}). The facial landmarks model had moderate predictive power with a test AUC of 0.67. Table \ref{tab:performance_metrics} summarizes the performance metrics for each model.
\begin{table*}

\centering
\resizebox{\columnwidth}{!}{%
\begin{tabular}{|l|c|c|c|}
\hline
\textbf{Metric} & \textbf{Eye Gazing} & \textbf{Facial Landmarks} & \textbf{Head Pose} \\
\hline
\textbf{AUC score} & 0.86 [0.79, 0.92] & 0.67 [0.55, 0.78] & 0.78 [0.69, 0.86] \\
\hline
\textit{Accuracy} & 0.79 [0.72, 0.86] & 0.75 [0.68, 0.82] & 0.75 [0.68, 0.82] \\
\hline
\textit{Recall (MA)} & 0.84 [0.76, 0.91] & 0.65 [0.55, 0.76] & 0.65 [0.55, 0.76] \\
\textit{Recall (WA)} & 0.79 [0.72, 0.86] & 0.75 [0.68, 0.82] & 0.75 [0.68, 0.82] \\
\hline
\textit{Precision (MA)} & 0.72 [0.65, 0.79] & 0.62 [0.53, 0.70] & 0.62 [0.53, 0.70] \\
\textit{Precision (WA)} & 0.89 [0.84, 0.92] & 0.79 [0.71, 0.86] & 0.79 [0.71, 0.86] \\
\hline
\textbf{F1-score (MA)} & 0.73 [0.65, 0.81] & 0.63 [0.53, 0.71] & 0.63 [0.53, 0.71] \\
\textit{F1-score (WA)} & 0.82 [0.75, 0.87] & 0.77 [0.69, 0.83] & 0.77 [0.69, 0.83] \\
\hline
\end{tabular}%
}
\caption{Performance Metrics for the Best Model of Each Task. The default threshold chosen for classification was 0.5.}
\label{tab:performance_metrics}

\end{table*}

The confidence intervals in Table \ref{tab:performance_metrics} were obtained using bootstrapping. We resampled the test set 1000 times, calculating each sample's metrics. The 2.5th and 97.5th percentiles of these metrics provide a 95\% confidence interval.

The eye model outperforms the facial landmarks and head pose models, with narrower confidence intervals indicating consistent performance. The eye model's high F1-score reflects its balanced precision and recall, demonstrating the best overall performance.

Combining the three modalities enhanced predictive power. The late fusion models, particularly the averaging method (test AUC of 0.90) and the linear method (test AUC of 0.84), performed strongly. The intermediate fusion model performed poorly (test AUC of 0.55).

We also tested Two-by-two feature combinations. For late fusion by averaging, the Eye+Head model achieved an AUC of 0.87, Face+Head 0.78, and Eye+Face 0.87. For late fusion with a linear layer, Eye+Head achieved an AUC of 0.82, Face+Head 0.67, and Eye+Face 0.90. The Eye+Face combination excelled in both methods, leveraging eye gaze and facial features for robust predictions.

Table \ref{tab:fusion_performance_metrics} summarizes the best fusion models' performance metrics. The late fusion models outperformed the intermediate fusion model. The late fusion-averaging model showed the highest AUC and balanced precision and recall. The intermediate fusion model had low AUC and precision, indicating inadequacy. Narrower confidence intervals for the late fusion models suggest consistent performance and lower variability. The averaging model's high macro-averaged F1-score indicates the best overall performance.

\begin{table*}[h]
\centering
\resizebox{\columnwidth}{!}{%
\begin{tabular}{|l|c|c|c|c|}
\hline
\textbf{Metric} & \textbf{\makecell{Late Fusion\\ (Eye, Head, Face)\\ Averaging}} & \textbf{\makecell{Late Fusion\\ (Eye, Head, Face)\\ Linear}} & \textbf{\makecell{Late Fusion\\ (Eye, Face)\\ Linear}} \\
\hline
\textbf{AUC score} & 0.90 [0.84, 0.95] & 0.84 [0.77, 0.91] & 0.90 [0.83, 0.95] \\
\hline
{Accuracy} & 0.82 [0.75, 0.89] & 0.84 [0.78, 0.91] & 0.89 [0.83, 0.93] \\
\hline
{Recall (MA)} & 0.88 [0.81, 0.92] & 0.89 [0.82, 0.94] & 0.85 [0.76, 0.93] \\
{Recall (WA)} & 0.82 [0.75, 0.89] & 0.84 [0.78, 0.91] & 0.89 [0.84, 0.93] \\
\hline
{Precision (MA)} & 0.75 [0.67, 0.82] & 0.76 [0.69, 0.85] & 0.80 [0.72, 0.89] \\
{Precision (WA)} & 0.90 [0.87, 0.93] & 0.91 [0.88, 0.94] & 0.90 [0.85, 0.94] \\
\hline
\textbf{F1-score (MA)} & 0.77 [0.69, 0.85] & 0.79 [0.71, 0.88] & 0.82 [0.74, 0.90] \\
{F1-score (WA)} & 0.84 [0.78, 0.90] & 0.86 [0.80, 0.92] & 0.89 [0.83, 0.93] \\
\hline
\end{tabular}%
}
\caption{Performance Metrics for the Best Fusion Models. The 95\% intervals were obtained by bootstrapping the test set. The default threshold chosen for classification was 0.5.}
\label{tab:fusion_performance_metrics}
\end{table*}

\subsection{Fairness evaluation}

We evaluated our models for age and gender sensitivity, excluding geographic locations due to insufficient demographic parity differences and equalized odds differences, where a lower parity difference indicates more evenly distributed positive outcomes across groups, while a lower equalized odds difference indicates more evenly distributed error rates.

The Eye Model performs well \ref{tab:fairness}for age groups 1-4 and 9-12 but struggles with age group 5-8, showing moderate fairness issues (Demographic Parity Difference: 0.1732, Equalized Odds Difference: 0.1569). The Face and Head Models perform well for age group 1-4 but poorly for age group 9-12, with significant fairness challenges (Equalized Odds Difference: 0.7460). The Late Fusion (Avg) model shows improved performance and fairness across all age groups (Demographic Parity Difference: 0.1826, Equalized Odds Difference: 0.1250).
\begin{table*}[htbp]
\centering
\resizebox{\columnwidth}{!}{%
\begin{tabular}{|l|l|l|l|l|l|l|l|}
\hline
\multicolumn{1}{|c|}{\textbf{Model}} & \multicolumn{1}{c|}{\textbf{Accuracy}} & \multicolumn{1}{c|}{\textbf{Recall}} & \multicolumn{1}{c|}{\textbf{Precision}} & \multicolumn{1}{c|}{\textbf{ROC AUC}} & \multicolumn{1}{c|}{\textbf{F1 Score}} & \multicolumn{1}{c|}{\textbf{\makecell{Demographic\\ Parity\\ Difference}}} & \multicolumn{1}{c|}{\textbf{\makecell{Equalized\\ Odds\\ Difference}}} \\ \hline
\textbf{Eye Model} & \cellcolor{black!55} & \cellcolor{black!55} & \cellcolor{black!55} & \cellcolor{black!55} & \cellcolor{black!55} & \cellcolor{black!55} & \cellcolor{black!55} \\ \hline
1-4 & 0.8276 & 0.8235 & 0.9767 & 0.8403 & 0.8936 & & \\ \hline
5-8 & 0.7045 & 0.6667 & 0.9600 & 0.7708 & 0.7869 & \cellcolor{white}0.1732 & \cellcolor{white}0.1569 \\ \hline
9-12 & 0.8529 & 0.8000 & 1.0000 & 0.9000 & 0.8889 & & \\ \hline
\textbf{Face Model} & \cellcolor{black!55} & \cellcolor{black!55} & \cellcolor{black!55} & \cellcolor{black!55} & \cellcolor{black!55} & \cellcolor{black!55} & \cellcolor{black!55} \\ \hline
1-4 & 0.8276 & 0.8235 & 0.9767 & 0.8403 & 0.8936 & & \\ \hline
5-8 & 0.7273 & 0.7500 & 0.9000 & 0.6875 & 0.8182 & \cellcolor{white}0.1711 & \cellcolor{white}0.7460 \\ \hline
9-12 & 0.6471 & 0.8400 & 0.7241 & 0.4756 & 0.7778 & & \\ \hline
\textbf{Head Model} & \cellcolor{black!55} & \cellcolor{black!55} & \cellcolor{black!55} & \cellcolor{black!55} & \cellcolor{black!55} & \cellcolor{black!55} & \cellcolor{black!55} \\ \hline
1-4 & 0.8276 & 0.8235 & 0.9767 & 0.8403 & 0.8936 & & \\ \hline
5-8 & 0.7273 & 0.7500 & 0.9000 & 0.6875 & 0.8182 & \cellcolor{white}0.1711 & \cellcolor{white}0.7460 \\ \hline
9-12 & 0.6471 & 0.8400 & 0.7241 & 0.4756 & 0.7778 & & \\ \hline
\textbf{Late Fusion (Avg)} & \cellcolor{black!55} & \cellcolor{black!55} & \cellcolor{black!55} & \cellcolor{black!55} & \cellcolor{black!55} & \cellcolor{black!55} & \cellcolor{black!55} \\ \hline
1-4 & 0.8621 & 0.8431 & 1.0000 & 0.9216 & 0.9149 & & \\ \hline
5-8 & 0.7727 & 0.7500 & 0.9643 & 0.8125 & 0.8438 & \cellcolor{white}0.1826 & \cellcolor{white}0.1250 \\ \hline
9-12 & 0.8235 & 0.7600 & 1.0000 & 0.8800 & 0.8636 & & \\ \hline
\textbf{Late Fusion (Linear)} & \cellcolor{black!55} & \cellcolor{black!55} & \cellcolor{black!55} & \cellcolor{black!55} & \cellcolor{black!55} & \cellcolor{black!55} & \cellcolor{black!55} \\ \hline
1-4 & 0.8621 & 0.8431 & 1.0000 & 0.9216 & 0.9149 & & \\ \hline
5-8 & 0.8182 & 0.8056 & 0.9667 & 0.8403 & 0.8788 & \cellcolor{white}0.1531 & \cellcolor{white}0.1250 \\ \hline
9-12 & 0.8529 & 0.8000 & 1.0000 & 0.9000 & 0.8889 & & \\ \hline
\textbf{Late Fusion (Eye+Face)} & \cellcolor{black!55} & \cellcolor{black!55} & \cellcolor{black!55} & \cellcolor{black!55} & \cellcolor{black!55} & \cellcolor{black!55} & \cellcolor{black!55} \\ \hline
1-4 & 0.9655 & 0.9608 & 1.0000 & 0.9804 & 0.9800 & & \\ \hline
5-8 & 0.8182 & 0.8611 & 0.9118 & 0.7431 & 0.8857 & \cellcolor{white}0.1389 & \cellcolor{white}0.3750 \\ \hline
9-12 & 0.8529 & 0.8800 & 0.9167 & 0.8289 & 0.8980 & & \\ \hline
\end{tabular}}%
\caption{Summary of Model Fairness across Age Groups.}
\label{tab:fairness}
\end{table*}

Regarding gender, the Eye Model performs slightly better for females \ref{tab:model_analysis_gender}, with balanced fairness metrics (Demographic Parity Difference: 0.1078, Equalized Odds Difference: 0.0268). The Face and Head Models show more gender disparities, with lower performance for females and higher fairness differences (Demographic Parity Difference: 0.2888, Equalized Odds Difference: 0.4196). The Late Fusion (Avg) model improves gender performance and fairness (Demographic Parity Difference: 0.2071, Equalized Odds Difference: 0.0769), while the Late Fusion (Linear) model offers the best balance of performance and fairness (Demographic Parity Difference: 0.1979, Equalized Odds Difference: 0.0769).
\begin{table*}[h]
    \centering
    \resizebox{\columnwidth}{!}{%
    \begin{tabular}{|l|l|l|l|l|l|l|l|}
        \hline
        \multicolumn{1}{|c|}{\textbf{Model}} & \multicolumn{1}{c|}{\textbf{Accuracy}} & \multicolumn{1}{c|}{\textbf{Recall}} & \multicolumn{1}{c|}{\textbf{Precision}} & \multicolumn{1}{c|}{\textbf{ROC AUC}} & \multicolumn{1}{c|}{\textbf{F1 Score}} & \multicolumn{1}{c|}{\textbf{\makecell{Demographic\\ Parity\\ Difference}}} & \multicolumn{1}{c|}{\textbf{\makecell{Equalized\\ Odds\\ Difference}}} \\ \hline
        \textbf{Eye Model}                 & \cellcolor{black!55} & \cellcolor{black!55} & \cellcolor{black!55} & \cellcolor{black!55} & \cellcolor{black!55} & \cellcolor{black!55} & \cellcolor{black!55} \\ \hline
        Female                               & 0.8235                                 & 0.7826                                & 0.9474                                  & 0.8458                                & 0.8571                                 &                                                         &                                                        \\ \hline
        Male                                 & 0.7778                                 & 0.7558                                & 0.9848                                  & 0.8394                                & 0.8553                                 & \cellcolor{white}0.1078 & \cellcolor{white}0.0268 \\ \hline
        \textbf{Face Model}                & \cellcolor{black!55} & \cellcolor{black!55} & \cellcolor{black!55} & \cellcolor{black!55} & \cellcolor{black!55} & \cellcolor{black!55} & \cellcolor{black!55} \\ \hline
        Female                               & 0.6765                                 & 0.6522                                & 0.8333                                  & 0.6897                                & 0.7317                                 &                                                         &                                                        \\ \hline
        Male                                 & 0.7677                                 & 0.8372                                & 0.8889                                  & 0.5725                                & 0.8623                                 & \cellcolor{white}0.2888 & \cellcolor{white}0.4196 \\ \hline
        \textbf{Head Model}                & \cellcolor{black!55} & \cellcolor{black!55} & \cellcolor{black!55} & \cellcolor{black!55} & \cellcolor{black!55} & \cellcolor{black!55} & \cellcolor{black!55} \\ \hline
        Female                               & 0.6765                                 & 0.6522                                & 0.8333                                  & 0.6897                                & 0.7317                                 &                                                         &                                                        \\ \hline
        Male                                 & 0.7677                                 & 0.8372                                & 0.8889                                  & 0.5725                                & 0.8623                                 & \cellcolor{white}0.2888 & \cellcolor{white}0.4196 \\ \hline
        \textbf{Late Fusion (Avg)}          & \cellcolor{black!55} & \cellcolor{black!55} & \cellcolor{black!55} & \cellcolor{black!55} & \cellcolor{black!55} & \cellcolor{black!55} & \cellcolor{black!55} \\ \hline
        Female                               & 0.8235                                 & 0.7391                                & 1.0000                                  & 0.8696                                & 0.8500                                 &                                                         &                                                        \\ \hline
        Male                                 & 0.8182                                 & 0.8023                                & 0.9857                                  & 0.8627                                & 0.8846                                 & \cellcolor{white}0.2071 & \cellcolor{white}0.0769 \\ \hline
        \textbf{Late Fusion (Linear)}       & \cellcolor{black!55} & \cellcolor{black!55} & \cellcolor{black!55} & \cellcolor{black!55} & \cellcolor{black!55} & \cellcolor{black!55} & \cellcolor{black!55} \\ \hline
        Female                               & 0.8529                                 & 0.7826                                & 1.0000                                  & 0.8913                                & 0.8780                                 &                                                         &                                                        \\ \hline
        Male                                 & 0.8384                                 & 0.8256                                & 0.9861                                  & 0.8743                                & 0.8987                                 & \cellcolor{white}0.1979 & \cellcolor{white}0.0769 \\ \hline
        \textbf{Late Fusion (Eye+Face)} & \cellcolor{black!55} & \cellcolor{black!55} & \cellcolor{black!55} & \cellcolor{black!55} & \cellcolor{black!55} & \cellcolor{black!55} & \cellcolor{black!55} \\ \hline
    Female & 0.8235 & 0.8261 & 0.9048 & 0.8221 & 0.8636 & & \\ \hline
    Male & 0.9091 & 0.9302 & 0.9639 & 0.8497 & 0.9467 & \cellcolor{white}0.2207 & \cellcolor{white}0.1041 \\ \hline
    \end{tabular}}%
\caption{Summary of Model Fairness concerning Gender. The class Other/None is ignored since we only have a sample size of 3 in the test set which is not representative.}
    \label{tab:model_analysis_gender}
\end{table*}
Given the importance of early diagnosis, the Late Fusion (Linear) model is the most balanced and fair option for both age and gender groups. Fairness mitigation techniques applied to the Head and Face Models for the age group 9-12 led to marginal improvements, so we focused on the fairer and more effective Late Fusion Methods.

\subsection{Net Benefit Analysis}

\begin{figure}[h]
\centering
    \includegraphics[width=0.7\columnwidth]{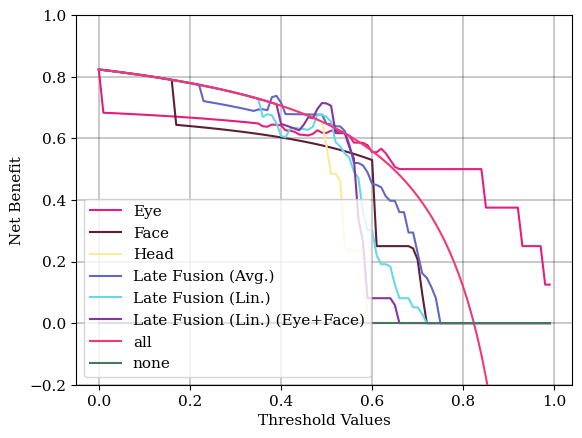}
    \caption{Net Benefit Analysis. Each line on the graph shows a model’s net benefit (y-axis) against threshold values (x-axis). The dark green line ("all") represents treating all as ASD, while the green line ("none") represents treating none, serving as baselines.}
    \label{fig:net_benefit_analysis}
\end{figure}

The Net Benefit Analysis curve (Figure \ref{fig:net_benefit_analysis}) compares different models across various thresholds, illustrating their performance in terms of net benefit. High sensitivity ensures most children with ASD are correctly diagnosed for timely interventions, while high specificity prevents incorrect diagnoses of NT children, avoiding unnecessary interventions. This balance ensures a reliable and practical diagnostic process, efficiently allocating resources.

%Each line on the graph represents a model's net benefit, with the y-axis showing net benefit and the x-axis indicating threshold values. The dark green line labeled ``all'' represents treating all individuals as having ASD, while the green line labeled ``none'' represents not treating any individuals with ASD, providing baseline comparisons. Compared to the baselines of treating all individuals as having ASD ("all") and treating all as not having ASD ("none"), TEST

%The Late Fusion (Avg) model consistently offers a higher net benefit across thresholds compared to other models. The Late Fusion (Linear) model also performs well but less consistently. The Eye Model has a good net benefit, lower than fusion models and the ``all'' strategy at smaller thresholds but higher at thresholds above 0.6. The Face and Head Models exhibit lower net benefits, with the Head Model showing significant drops at certain thresholds.

%The ``none'' strategy has lower net benefits compared to optimized models, while the ``all'' strategy has higher net benefits only at lower thresholds. This analysis doesn't reflect the actual distribution of ASD and NT children in the general population, as the clinical trial dataset is skewed with more ASD children than NT. In reality, ASD affects about 1 in 36 children. This discrepancy affects net benefit analysis, as the sample isn't representative of the general population, impacting the application of these models in real-world contexts.
 The Late Fusion (Avg) model consistently offers higher net benefits, with the Late Fusion (Linear) and Eye Model performing well at varying thresholds. The Face and Head Models show lower net benefits. The dataset's skewed ASD prevalence affects the generalizability of these findings.

\section{Limitations and Future Directions}

%\subsection{Limitations and Future Directions}
%To conclude, we developed a protocol to filter high-quality videos from the GuessWhat dataset and compiled a minimally viable training set. Based on this training set, we constructed individual LSTMs-based models using eye gazing, head positions, and facial landmarks as input features to predict ASD, achieving a test AUC score of approximately 80\%. Finally, we implemented both late fusion and intermediate fusion techniques to build ensemble models, further enhancing predictive performance to 90 \% and producing a fairer model across gender and age groups (late fusion (Linear) model). 

%\subsection{Limitations and Future Directions}
Our current models face several limitations. First, incorporating accelerometer data can mitigate data drifts, ensuring more reliable predictions. Secondly, developing an automated preprocessing pipeline is crucial for handling challenges such as detecting when a child is too far or close to the camera, managing multiple faces, and tracking different children in videos. As we continue to collect more data, an age-centric approach will allow us to tailor games and features to different age groups, thereby enhancing both engagement and the informativeness of the features extracted. Moreover, expanding our models to incorporate additional modalities, such as speech, will provide us with a richer understanding of subjects. Additionally, focusing on time-series pre-training and improving interpretability will enhance robustness and transparency, making models more reliable and understandable. To ensure fairness, it is crucial to collect skin color data and generalize across different skin tones, thereby promoting equity and reducing bias in our predictive outcomes.

%\subsection{Acknoweledgments}
%\subsection{Broader Impacts}
%This work contributes to the early detection of ASD, reduces reliance on subjective assessment of ASD, and improves health outcomes for children with ASD. Moreover, this approach can be translated to other diseases and diagnostic procedures, such as ADHD, and digital phenotyping, broadening its impact on healthcare.

%%%%%%%%% REFERENCES
{\small
\bibliographystyle{ieee_fullname}
\bibliography{egbib}
}

\newpage
\onecolumn
\appendix
\section{Appendix}

 All thresholds are summarized in Table~\ref{tab:thresholds}
\begin{table*}[h]
\centering
\begin{tabular}{ccc}
\toprule
\textbf{\makecell{Filtering Criteria}} & \textbf{\makecell{Variables}} & \textbf{\makecell{Thresholds}} \\
\midrule
\makecell{Quality} & \makecell{Sharpness\\ Brightness} 
& \makecell{Sharpness (from 0 to 100) $>$ 4\\ 
Brightness (From 0 to 100) $>$ 20} \\
\midrule
\makecell{Face Detection} & \makecell{No Face Proportion\\ MultiFace Proportion\\ Face Size} & \makecell{No Face Detection (0 to 1) $<$ 0.6\\ MultiFace Detection (0 to 1) $<$ 0.3\\ Face Size (0 to 100) $>$ 0.01} \\
\midrule
\makecell{Eye Visibility} & \makecell{Head Pose (Pitch, Roll, Yaw)\\ Eyes Confidence} & \makecell{Pitch (from -180 to 180) $<$ 45\\ Roll (from -180 to 180) $<$ 45\\ Yaw (from -180 to 180) $<$ 45\\ Eye Confidence (0 to 100) $>$ 75} \\
\bottomrule
\end{tabular}
\caption{Filtering Thresholds based on Features Extracted using AWS \cite{awsrekognition2023} for each video. These criteria constitute the conditions for a video to be considered.}
\label{tab:thresholds}  
%Add a label for referencing this table in your text
\end{table*}

\begin{table}[h]
\centering
\caption{Hyperparameter Search Space for the Individual Models}
\begin{tabular}{@{}lc@{}}
\toprule
Hyperparameter & Range Searched \\ 
\midrule
Model & \makecell{\{LSTM, GRU, CNN+LSTM, CNN+GRU\}} \\
\makecell[l]{Hidden size\\ (of LSTM/GRU)} & \{16, 32, 64\}\\
Batch Size & \{32, 48, 64, 100\} \\
Num of Layers & [4, 8]\\
Dropout Probability & [0.1, 0.3]\\
Learning Rate & [1e-4, 1e-1] \\ 
Weight Decay & [1e-5, 1e-2]\\
Optimizer & Adam \\
Loss Function & Cross-Entropy, Focal Loss \\
\bottomrule
\end{tabular}
\label{tab:search_space_individual}
\end{table}

\begin{table}[h]
\centering
\caption{Hyperparameter Search Space for the Intermediate Fusion Models}
\begin{tabular}{@{}lc@{}}
\toprule
Hyperparameter & Range Searched \\ 
\midrule
Batch Size & \{16, 32, 64\} \\
Learning Rate  & [$1e-4, 1e-1$]\\
First Hidden Size & \{128, 192, 256\} \\
Second Hidden Size & \{32, 64, 128\}\\
Third Hidden Size & \{32, 64\}\\
Optimizer & Adam \\
Loss Function & Cross-Entropy \\
\bottomrule
\end{tabular}
\label{tab:search_space_interm}
\end{table}

\begin{table*}[h]
\caption{Best Hyperparameters for Different Tasks Chosen Based on Validation Loss.}
\centering
\begin{tabular}{|l|c|c|c|}
\hline
\textbf{Hyperparameter} & \textbf{Eye Gazing} & \textbf{Head Pose} & \textbf{Facial Landmarks} \\
\hline
Model & LSTM & LSTM & LSTM \\
Hidden Size & 64 & 32 & 48 \\
Batch Size & 64 & 48 & 48 \\
Num of Layers & 8 & 4 & 4 \\
Dropout Probability & 0.265894 & 0.194464 & 0.179506 \\
Learning Rate & 0.0324491 & 0.000336268 & 0.0464146 \\
Weight Decay & 1.15693e-05 & 3.82511e-05 & 3.66175e-05 \\
Optimizer & Adam & Adam & Adam \\
Loss Function & Cross-Entropy & Cross-Entropy & Cross-Entropy \\
\hline
\end{tabular}
\label{tab:hyperparameters}
\end{table*}

\begin{table}[htbp]
    \centering
    \begin{tabular}{|l|l|}
        \hline
        \textbf{Hyperparameter} & \textbf{Value} \\ \hline
        batch\_size & 16 \\ \hline
        learning\_rate & 0.07165411551018012 \\ \hline
        first\_hidden\_size & 256 \\ \hline
        second\_hidden\_size & 32 \\ \hline
        third\_hidden\_size & 64 \\ \hline
        num\_epochs & 15 \\ \hline
    \end{tabular}
    \caption{Hyperparameters for Intermediate Fusion}
    \label{tab:intermediate_fusion}
\end{table}

\begin{table}[htbp]
    \centering
    \begin{tabular}{|l|l|}
        \hline
        \textbf{Hyperparameter} & \textbf{Value} \\ \hline
        batch\_size & 32 \\ \hline
        learning\_rate & 0.0005439380832835521 \\ \hline
        num\_epochs & 11 \\ \hline
    \end{tabular}
    \caption{Hyperparameters for Late Fusion (Linear)}
    \label{tab:late_fusion_linear}
\end{table}

% \begin{algorithm}[h]
% \SetKwInOut{Input}{input}
% \SetKwInOut{Output}{output}
% \Input{ $[X^1, ..., X^m]$ with $X^i =[x^i_1, ..., x^i_d]$ where m denotes the number of frames and d the dimension of the feature set.}
% \Output{$[X^a, ..., X^b]$ with $X^i =[x^i_1, ..., x^i_d]$ where a denotes the first non-None index, b the last non-None index and d the dimension of the feature set.}
% \Fn{\FMain}{
%   $a \leftarrow -1$\;
%   $b \leftarrow -1$\;
%   \For{$i \leftarrow 1$ \KwTo $m$}{
%     \If{$X^i \neq \text{None}$}{
%       $a \leftarrow i$\;
%       \textbf{break}\;
%     }
%   }
%   \For{$i \leftarrow m$ \KwTo $1$ \textbf{step} $-1$}{
%     \If{$X^i \neq \text{None}$}{
%       $b \leftarrow i$\;
%       \textbf{break}\;
%     }
%   }
%   \eIf{$a \neq -1$ \textbf{and} $b \neq -1$}{
%     \KwRet{$[X^a, ..., X^b]$}\;
%   }{
%     \KwRet{$[]$} \tcp*{All frames are None}
%   }
% }
% \caption{Truncate Window}
% \label{alg:truncate}
% \end{algorithm}

\begin{algorithm}[h]
\SetKwInOut{Input}{Input}
\SetKwInOut{Output}{Output}

\Input{$[X^1, ..., X^m]$ with $X^i =[x^i_1, ..., x^i_d]$ where $m$ denotes the number of frames and $d$ the dimension of the feature set.}
\Output{$[X^a, ..., X^b]$ with $X^i =[x^i_1, ..., x^i_d]$ where $a$ denotes the first non-None index, $b$ the last non-None index, and $d$ the dimension of the feature set.}

\SetKwFunction{FMain}{TruncateWindow}
\SetKwProg{Fn}{Function}{:}{}
\Fn{\FMain}{
  $a \leftarrow -1$\;
  $b \leftarrow -1$\;
  \For{$i \leftarrow 1$ \KwTo $m$}{
    \If{$X^i \neq \text{None}$}{
      $a \leftarrow i$\;
      \textbf{break}\;
    }
  }
  \For{$i \leftarrow m$ \KwTo $1$ \textbf{step} $-1$}{
    \If{$X^i \neq \text{None}$}{
      $b \leftarrow i$\;
      \textbf{break}\;
    }
  }
  \eIf{$a \neq -1$ \textbf{and} $b \neq -1$}{
    \KwRet{$[X^a, ..., X^b]$}\;
  }{
    \KwRet{$[]$} \tcp*{All frames are None}
  }
}
\caption{Truncate Window}
\label{alg:truncate}
\end{algorithm}

\begin{algorithm}[h]
\SetKwInOut{Input}{input}
\SetKwInOut{Output}{output}
\Input{$[X^1, ..., X^m]$ with $X^i =[x^i_1, ..., x^i_d]$ where m denotes the number of frames and d the dimension of the feature set.}
\Input{$s$, the number of seconds for determining uninformative missingness (e.g., $s=2$).}
\Input{$fps$, frames per second of the video.}
\Output{List of continuous windows with no more than s seconds of no face detection where each window $W$ is a sublist $[X^a, ..., X^b]$ such that $X^i =[x^i_1, ..., x^i_d]$.}

\SetKwData{$MaxMissing$}{$MaxMissing$Frames}
\SetKwData{$CurrentWindow$}{$CurrentWindow$}
\SetKwData{$AllWindows$}{$AllWindows$}
\SetKwData{$CountMissing$}{$CountMissing$}

$MaxMissing$ $\leftarrow s \times fps$ \\ % Calculate max missing frames based on input seconds and fps
$CurrentWindow$ {$\leftarrow$ []} \\
$AllWindows$ {$\leftarrow$ []} \\
$CountMissing$ {$\leftarrow$ 0}\\

\BlankLine
\For{$i \leftarrow 1$ \KwTo $m$}{
    \eIf{$X^i \neq \text{[None, ..., None]}$}{
        \If{ $CountMissing$ $>$ $MaxMissing$}{
            \If{$CurrentWindow$ $\neq$ []}{
                $AllWindows$.append(\\\textbf{Truncate}($CurrentWindow$))\;
                $CurrentWindow$ $\leftarrow$ []\;
            }
            $CountMissing$ $\leftarrow$ 0\;
        }
        $CurrentWindow$.append($X^i$)\;
        $CountMissing$ $\leftarrow$ 0\;  % Reset missing counter on valid data
    }{
        $CountMissing$ $\leftarrow$ $CountMissing$ + 1\;
        \If{$CountMissing$ $\leq$ $MaxMissing$}{
            $CurrentWindow$.append($X^i$)\;
        }
    }
}

\If{$CurrentWindow$ $\neq$ []}{
    $AllWindows$.append(\textbf{Truncate}($CurrentWindow$))\;
}

\BlankLine
\KwRet{$AllWindows$}
\caption{Create Windows}
\label{alg:create_windows}
\end{algorithm}

\begin{algorithm}[h]
\SetKwInOut{Input}{input}
\SetKwInOut{Output}{output}
\Input{$windows$, list of continuous windows with no more than s seconds of no face detection where each window $W$ is a sublist $[X^a, ..., X^b]$ such that $X^i =[x^i_1, ..., x^i_d]$}
\Input{$s$, minimal number of seconds of features for a window to be considered}
\Input{$fps$, framerate per second of the video}
\Output{Concatenated list of informative windows $[X^a, ..., X^b]$ with $X^i =[x^i_1, ..., x^i_d]$ where d the dimension of the feature set.}
$ConcatenatedWindow$ $\leftarrow$ []\\
\For{window in $windows$}{
    if $(len(window) \geq (s \times fps))${
    $ConcatenatedWindow$.append(window)
    }
}
\KwRet{ConcatenatedWindow}
\caption{Concatenate Windows}
\label{alg:concatenate}
\end{algorithm}

\begin{figure*}
    \centering
    \resizebox{1\textwidth}{!}{\input{fusion}}    
    \caption{Late and Intermediate Fusion Schemes.}
    \label{fig:fusion}
\end{figure*}
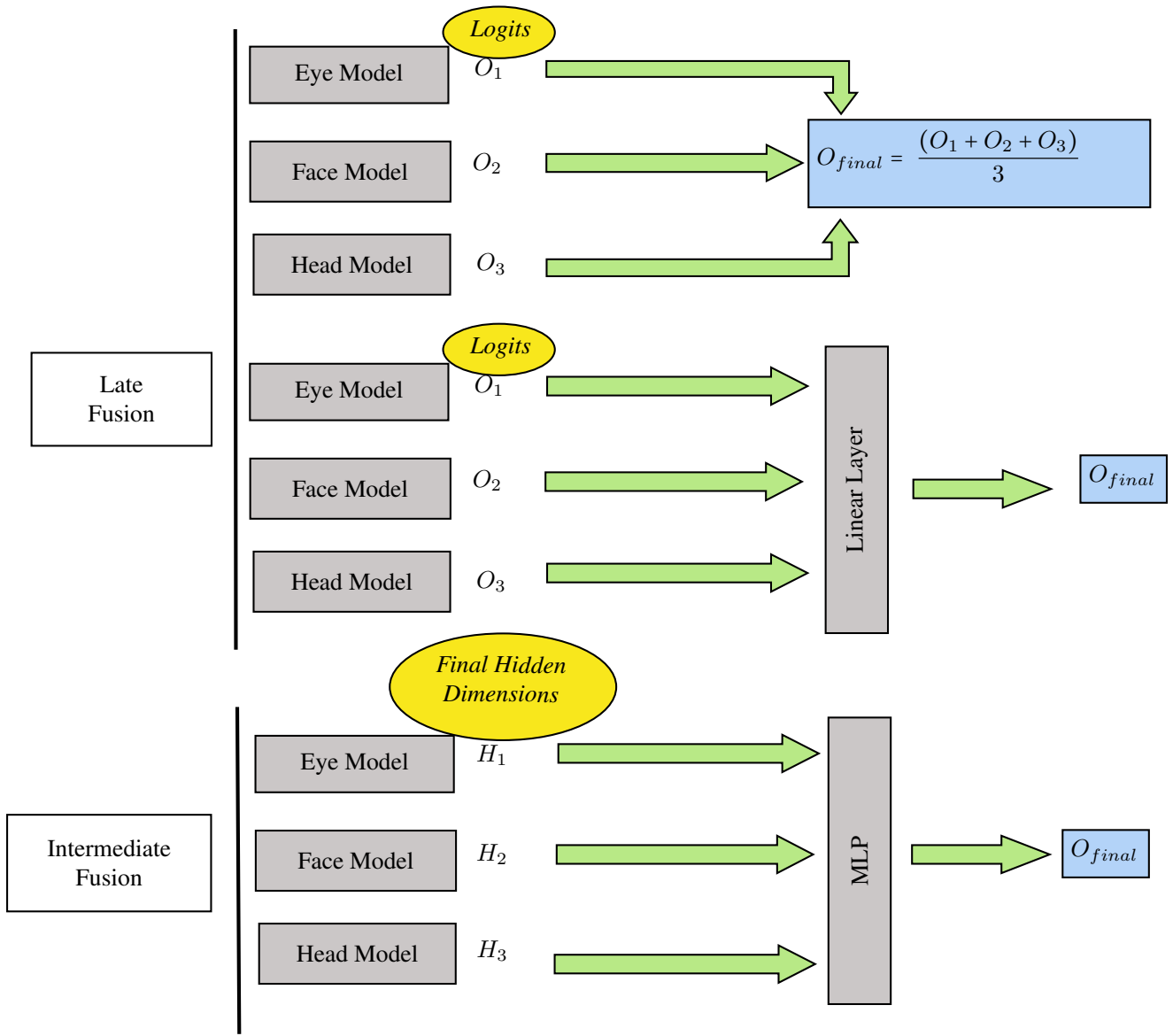

% \begin{figure*}[h]
%     \centering
%     \includegraphics[width=1\columnwidth]{latex/roc_curve_late_fusion_linear_combinations_version6.png}
%     \caption{Receiver Operation Characteristic Curves - Combinations (Linear).}
%     \label{fig:roc_curves_combinations}
% \end{figure*}

% \begin{figure*}[h]
%     \centering
%     \includegraphics[width=0.8\columnwidth]{latex/roc_curve_eye_face_head.png}
%     \caption{Receiver Operation Characteristic Curves (Individual Models).}
%     \label{fig:roc_curves_individuals}
% \end{figure*}

\end{document}

%% file: latex/figures/superusers_plot.tex
\begin{tikzpicture}

\definecolor{steelblue31119180}{RGB}{31,119,180}
\definecolor{darkorange25512714}{RGB}{255,127,14}

\begin{axis}[
    tick align=outside,
    tick pos=left,
    title={Number of Videos per Child},
    x grid style={white!69.0196078431373!black},
    xlabel={Number of Videos},
    xmin=-3.45, xmax=94.45,
    xtick style={color=black},
    y grid style={white!69.0196078431373!black},
    ylabel={Count (log scale)},
    ymin=0, ymax=172.2,
    ymode=log, % If you need log scale
    ytick style={color=black}
]

% ASD Histogram Data
\addplot[
    ybar,
    fill=steelblue31119180,
    draw=none,
    area legend
] table[row sep=crcr]{
    x y\\
    1 164\\
    5.45 22\\
    9.9 14\\
    14.35 12\\
    18.8 8\\
    23.25 5\\
    27.7 9\\
    32.15 2\\
    36.6 3\\
    41.05 1\\
    45.5 0\\
    49.95 1\\
    54.4 1\\
    58.85 1\\
    63.3 0\\
    67.75 0\\
    72.2 0\\
    76.65 0\\
    81.1 0\\
    85.55 1\\
    90 1\\
};
\addlegendentry{ASD}

% NT Histogram Data
\addplot[
    ybar,
    fill=darkorange25512714,
    draw=none,
    area legend
] table[row sep=crcr]{
    x y\\
    1 31\\
    2.2 4\\
    3.4 4\\
    4.6 0\\
    5.8 1\\
    7 1\\
    8.2 0\\
    9.4 1\\
    10.6 0\\
    11.8 0\\
    13 0\\
    14.2 0\\
    15.4 0\\
    16.6 0\\
    17.8 0\\
    19 0\\
    20.2 0\\
    21.4 0\\
    22.6 0\\
    23.8 1\\
};
\addlegendentry{NT}

\end{axis}
\end{tikzpicture}

%% file: latex/figures/filtering_steps.tex
\tikzset{every picture/.style={line width=0.75pt}} %set default line width to 0.75pt        

\begin{tikzpicture}[x=0.75pt,y=0.75pt,yscale=-1,xscale=1]
%uncomment if require: \path (0,358); %set diagram left start at 0, and has height of 358

%Rounded Rect [id:dp3515379037244586] 
\draw  [fill={rgb, 255:red, 74; green, 144; blue, 226 }  ,fill opacity=1 ] (10.2,79.92) .. controls (10.2,68.51) and (19.45,59.26) .. (30.86,59.26) -- (92.83,59.26) .. controls (104.24,59.26) and (113.49,68.51) .. (113.49,79.92) -- (113.49,160.2) .. controls (113.49,171.61) and (104.24,180.86) .. (92.83,180.86) -- (30.86,180.86) .. controls (19.45,180.86) and (10.2,171.61) .. (10.2,160.2) -- cycle ;
%Image [id:dp06825064473725484] 
\draw (59.97,141.24) node  {\includegraphics[width=54.34pt,height=42.94pt]{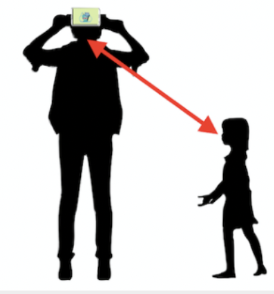}};
%Rounded Rect [id:dp16929281515187844] 
\draw  [fill={rgb, 255:red, 106; green, 165; blue, 238 }  ,fill opacity=1 ] (126.38,52.65) .. controls (126.38,42.62) and (141,20) .. (159.03,20) -- (262.48,20) .. controls (280.51,20) and (295.12,34.62) .. (295.12,52.65) -- (295.12,150.59) .. controls (295.12,168.62) and (280.51,183.23) .. (262.48,183.23) -- (159.03,183.23) .. controls (141,183.23) and (126.38,168.62) .. (126.38,150.59) -- cycle ;
%Rounded Rect [id:dp7668984541463255] 
\draw  [fill={rgb, 255:red, 158; green, 199; blue, 239 }  ,fill opacity=1 ] (307.68,52.65) .. controls (307.68,34.62) and (322.3,20) .. (340.33,20) -- (441.43,20) .. controls (459.46,20) and (474.08,34.62) .. (474.08,52.65) -- (474.08,150.59) .. controls (474.08,168.62) and (459.46,183.23) .. (441.43,183.23) -- (340.33,183.23) .. controls (322.3,183.23) and (307.68,168.62) .. (307.68,150.59) -- cycle ;
%Rounded Rect [id:dp2077557473653513] 
\draw  [fill={rgb, 255:red, 221; green, 235; blue, 255 }  ,fill opacity=1 ] (483.92,52.65) .. controls (483.92,34.62) and (498.54,20) .. (516.57,20) -- (619.55,20) .. controls (637.58,20) and (652.2,34.62) .. (652.2,52.65) -- (652.2,150.59) .. controls (652.2,168.62) and (637.58,183.23) .. (619.55,183.23) -- (516.57,183.23) .. controls (498.54,183.23) and (483.92,168.62) .. (483.92,150.59) -- cycle ;

% Text Node
\draw (10.85,67.25) node [anchor=north west][inner sep=0.75pt]  [font=\large] [align=left] {\begin{minipage}[lt]{70.97pt}\setlength\topsep{0pt}
\begin{center}
\textit{\textbf{\underline{Raw Videos}}}\\\textit{\textbf{\underline{(N = 3113)}}}
\end{center}

\end{minipage}};
% Text Node
\draw (161.03,23) node [anchor=north west][inner sep=0.75pt]   [align=left] {\begin{minipage}[lt]{60.89pt}\setlength\topsep{0pt}
\begin{center}
\textit{\textbf{\underline{{\large Dataset A}}}}\\\textit{{\large \textbf{\underline{(N = 2123}})}}
\end{center}

\end{minipage}};
% Text Node
\draw (347.48,22.72) node [anchor=north west][inner sep=0.75pt]  [font=\large] [align=left] {\textit{\textbf{\underline{Dataset B}}}\\\textit{\textbf{\underline{(N = 700}})}};
% Text Node
\draw (528.44,22.9) node [anchor=north west][inner sep=0.75pt]  [font=\large] [align=left] {\textit{\textbf{\underline{Dataset C}}}\\\textit{\textbf{\underline{(N = 688)}}}};
% Text Node
\draw (133.49,79.52) node [anchor=north west][inner sep=0.75pt]   [align=left] {\begin{minipage}[lt]{108.6pt}\setlength\topsep{0pt}
\begin{center}
{\small ASD/NT videos = \textbf{17:1}}\\{\small Male/Female videos = \textbf{3:1}}\\{\small \textbf{245} children with ASD}\\{\small \textbf{43} NT children}
\end{center}

\end{minipage}};
% Text Node
\draw (312.78,79.7) node [anchor=north west][inner sep=0.75pt]   [align=left] {\begin{minipage}[lt]{108.6pt}\setlength\topsep{0pt}
\begin{center}
{\small ASD/NT videos = \textbf{5:1}}\\{\small Male/Female videos = \textbf{3:1}}\\{\small \textbf{245} children with ASD}\\{\small \textbf{43} NT children}
\end{center}

\end{minipage}};
% Text Node
\draw (489.73,79.7) node [anchor=north west][inner sep=0.75pt]   [align=left] {\begin{minipage}[lt]{108.6pt}\setlength\topsep{0pt}
\begin{center}
{\small ASD/NT videos = \textbf{5:1}}\\{\small Male/Female videos = \textbf{3:1}}\\{\small \textbf{243} children with ASD}\\{\small \textbf{42} NT children}
\end{center}

\end{minipage}};
% Text Node
\draw  [fill={rgb, 255:red, 184; green, 233; blue, 134 }  ,fill opacity=1 ]  (124.75, 209.76) circle [x radius= 60.81, y radius= 59.4]   ;
\draw (82.75,169.76) node [anchor=north west][inner sep=0.75pt]   [align=left] {\begin{minipage}[lt]{59.02pt}\setlength\topsep{0pt}
\begin{center}
{\small \textit{Video Quality,}}\\{\small \textit{\% of Face,}}\\{\small \textit{Face Size,}}\\{\small \textit{Head Pose}}
\end{center}

\end{minipage}};
% Text Node
\draw  [fill={rgb, 255:red, 184; green, 233; blue, 134 }  ,fill opacity=1 ]  (300.34, 210.76) circle [x radius= 54.45, y radius= 59.4]   ;
\draw (262.84,170.76) node [anchor=north west][inner sep=0.75pt]   [align=left] {\begin{minipage}[lt]{52.56pt}\setlength\topsep{0pt}
\begin{center}
{\small \textit{Balance }}\\{\small \textit{ASD/NT,}}\\{\small \textit{Balance }}\\{\small \textit{Super-users}}
\end{center}

\end{minipage}};
% Text Node
\draw  [fill={rgb, 255:red, 184; green, 233; blue, 134 }  ,fill opacity=1 ]  (480.51, 210.57) circle [x radius= 84.85, y radius= 59.4]   ;
\draw (421.51,170.57) node [anchor=north west][inner sep=0.75pt]   [align=left] {\begin{minipage}[lt]{81.83pt}\setlength\topsep{0pt}
\begin{center}
{\small Discard videos}\\{\small with unsufficient }\\{\small features after}\\{\small feature engineering}
\end{center}

\end{minipage}};

\end{tikzpicture}

%% file: latex/figures/feature_extraction.tex
\tikzset{every picture/.style={line width=0.75pt}} %set default line width to 0.75pt        

\begin{tikzpicture}[x=0.75pt,y=0.75pt,yscale=-1,xscale=1]
%uncomment if require: \path (0,773); %set diagram left start at 0, and has height of 773

%Down Arrow [id:dp7022736389909261] 
\draw  [fill={rgb, 255:red, 155; green, 155; blue, 155 }  ,fill opacity=1 ] (214.47,120.91) -- (217.9,120.91) -- (217.9,110.34) -- (224.78,110.34) -- (224.78,120.91) -- (228.21,120.91) -- (221.34,127.96) -- cycle ;
%Down Arrow [id:dp9955880036882536] 
\draw  [fill={rgb, 255:red, 155; green, 155; blue, 155 }  ,fill opacity=1 ] (330.91,120.91) -- (334.34,120.91) -- (334.34,110.34) -- (341.22,110.34) -- (341.22,120.91) -- (344.66,120.91) -- (337.78,127.96) -- cycle ;
%Down Arrow [id:dp8644206004035189] 
\draw  [fill={rgb, 255:red, 155; green, 155; blue, 155 }  ,fill opacity=1 ] (487.75,120.91) -- (491.19,120.91) -- (491.19,110.34) -- (498.06,110.34) -- (498.06,120.91) -- (501.5,120.91) -- (494.63,127.96) -- cycle ;
%Down Arrow [id:dp6324937951276093] 
\draw  [fill={rgb, 255:red, 155; green, 155; blue, 155 }  ,fill opacity=1 ] (602.96,120.91) -- (606.4,120.91) -- (606.4,110.34) -- (613.27,110.34) -- (613.27,120.91) -- (616.71,120.91) -- (609.84,127.96) -- cycle ;
%Image [id:dp5196186120883854] 
\draw (492.44,73.83) node  {\includegraphics[width=57.29pt,height=43.62pt]{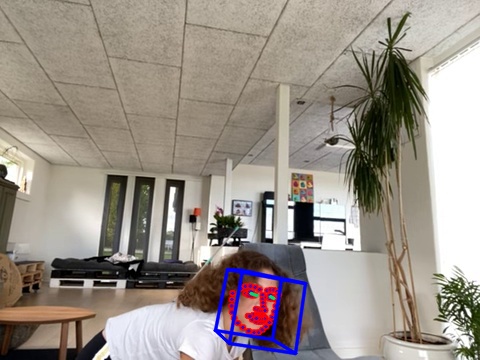}};
%Image [id:dp28667861218205015] 
\draw (220.03,73.47) node  {\includegraphics[width=57.46pt,height=43.74pt]{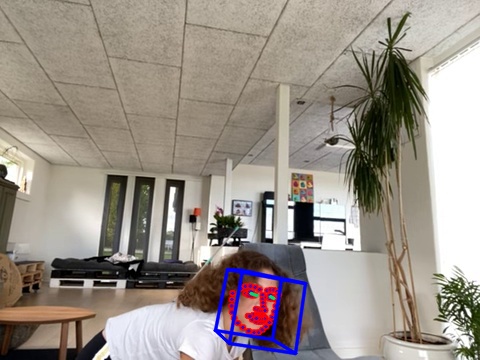}};
%Image [id:dp39596990017220923] 
\draw (337.14,73.47) node  {\includegraphics[width=57.46pt,height=43.74pt]{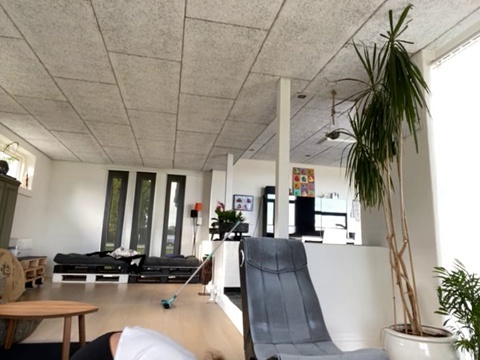}};
%Image [id:dp03545269437191245] 
\draw (607.36,73.83) node  {\includegraphics[width=57.29pt,height=43.62pt]{latex/figures/frame0064.jpg}};
%Shape: Rectangle [id:dp17695844576405362] 
\draw  [fill={rgb, 255:red, 184; green, 233; blue, 134 }  ,fill opacity=1 ] (163.97,455.66) -- (657.56,455.66) -- (657.56,510.02) -- (163.97,510.02) -- cycle ;
%Shape: Rectangle [id:dp16233736278328248] 
\draw  [fill={rgb, 255:red, 184; green, 233; blue, 134 }  ,fill opacity=1 ] (154.14,469.8) -- (651.56,469.8) -- (651.56,528.4) -- (154.14,528.4) -- cycle ;
%Shape: Rectangle [id:dp9579284477041377] 
\draw  [fill={rgb, 255:red, 184; green, 233; blue, 134 }  ,fill opacity=1 ] (148.2,483.8) -- (643.56,483.8) -- (643.56,542.4) -- (148.2,542.4) -- cycle ;

% Text Node
\draw  [fill={rgb, 255:red, 184; green, 233; blue, 134 }  ,fill opacity=1 ]  (172.86,5.66) -- (655.86,5.66) -- (655.86,36.66) -- (172.86,36.66) -- cycle  ;
\draw (414.36,21.16) node   [align=left] {\begin{minipage}[lt]{326.54pt}\setlength\topsep{0pt}
\begin{center}
{\large Video (90 seconds)}
\end{center}

\end{minipage}};
% Text Node
\draw (408.23,68.31) node [anchor=north west][inner sep=0.75pt]   [align=left] {\textbf{...}};
% Text Node
\draw  [fill={rgb, 255:red, 203; green, 199; blue, 199 }  ,fill opacity=1 ]  (1.47,61.83) -- (136.47,61.83) -- (136.47,92.83) -- (1.47,92.83) -- cycle  ;
\draw (68.97,77.33) node   [align=left] {\begin{minipage}[lt]{89.79pt}\setlength\topsep{0pt}
\begin{center}
Frames (10 fps)
\end{center}

\end{minipage}};
% Text Node
\draw  [fill={rgb, 255:red, 203; green, 199; blue, 199 }  ,fill opacity=1 ]  (1.47,148.56) -- (136.47,148.56) -- (136.47,214.56) -- (1.47,214.56) -- cycle  ;
\draw (68.97,181.56) node   [align=left] {\begin{minipage}[lt]{89.79pt}\setlength\topsep{0pt}
\begin{center}
Eye Gazing\\(AWS Rekognition)
\end{center}

\end{minipage}};
% Text Node
\draw  [fill={rgb, 255:red, 248; green, 231; blue, 28 }  ,fill opacity=1 ]  (174.86,140.81) -- (270.86,140.81) -- (270.86,226.81) -- (174.86,226.81) -- cycle  ;
\draw (222.86,183.81) node   [align=left] {\begin{minipage}[lt]{63.29pt}\setlength\topsep{0pt}
\begin{center}
\textbf{2 features}\\\textit{{\small Yaw,}}\\{\small \textit{Pitch}}\\
\end{center}

\end{minipage}};
% Text Node
\draw (408.01,168.57) node [anchor=north west][inner sep=0.75pt]   [align=left] {\textbf{...}};
% Text Node
\draw  [fill={rgb, 255:red, 203; green, 199; blue, 199 }  ,fill opacity=1 ]  (2.52,259.87) -- (136.52,259.87) -- (136.52,325.87) -- (2.52,325.87) -- cycle  ;
\draw (69.52,292.87) node   [align=left] {\begin{minipage}[lt]{89.05pt}\setlength\topsep{0pt}
\begin{center}
Face Landmarks\\(AWS Rekognition)
\end{center}

\end{minipage}};
% Text Node
\draw  [fill={rgb, 255:red, 203; green, 199; blue, 199 }  ,fill opacity=1 ]  (2.61,369.41) -- (138.61,369.41) -- (138.61,438.41) -- (2.61,438.41) -- cycle  ;
\draw (70.61,403.91) node   [align=left] {\begin{minipage}[lt]{90.54pt}\setlength\topsep{0pt}
\begin{center}
Head Landmarks\\(AWS \\Rekognition)
\end{center}

\end{minipage}};
% Text Node
\draw  [fill={rgb, 255:red, 233; green, 166; blue, 174 }  ,fill opacity=1 ]  (290.15,140.81) -- (386.15,140.81) -- (386.15,226.81) -- (290.15,226.81) -- cycle  ;
\draw (338.15,183.81) node   [align=left] {\begin{minipage}[lt]{63.2pt}\setlength\topsep{0pt}
\begin{center}
No features Detected
\end{center}

\end{minipage}};
% Text Node
\draw  [fill={rgb, 255:red, 254; green, 243; blue, 191 }  ,fill opacity=1 ]  (445.05,140.81) -- (541.05,140.81) -- (541.05,226.81) -- (445.05,226.81) -- cycle  ;
\draw (493.05,183.81) node   [align=left] {\begin{minipage}[lt]{63.2pt}\setlength\topsep{0pt}
\begin{center}

\end{center}

\end{minipage}};
% Text Node
\draw  [fill={rgb, 255:red, 254; green, 243; blue, 191 }  ,fill opacity=1 ]  (561.5,140.81) -- (657.5,140.81) -- (657.5,226.81) -- (561.5,226.81) -- cycle  ;
\draw (609.5,183.81) node   [align=left] {\begin{minipage}[lt]{63.2pt}\setlength\topsep{0pt}
\begin{center}

\end{center}

\end{minipage}};
% Text Node
\draw  [fill={rgb, 255:red, 248; green, 231; blue, 28 }  ,fill opacity=1 ]  (174.91,253.05) -- (271.91,253.05) -- (271.91,339.05) -- (174.91,339.05) -- cycle  ;
\draw (223.41,296.05) node   [align=left] {\begin{minipage}[lt]{64.04pt}\setlength\topsep{0pt}
\begin{center}
\textbf{60 features}\\{\small \textit{Eyes, Nose, Mouth, Face Shape}}\\
\end{center}

\end{minipage}};
% Text Node
\draw (409.11,284.53) node [anchor=north west][inner sep=0.75pt]   [align=left] {\textbf{...}};
% Text Node
\draw  [fill={rgb, 255:red, 233; green, 166; blue, 174 }  ,fill opacity=1 ]  (290.15,253.05) -- (386.15,253.05) -- (386.15,339.05) -- (290.15,339.05) -- cycle  ;
\draw (338.15,296.05) node   [align=left] {\begin{minipage}[lt]{63.2pt}\setlength\topsep{0pt}
\begin{center}
No features Detected
\end{center}

\end{minipage}};
% Text Node
\draw  [fill={rgb, 255:red, 254; green, 243; blue, 191 }  ,fill opacity=1 ]  (445.05,253.05) -- (541.05,253.05) -- (541.05,339.05) -- (445.05,339.05) -- cycle  ;
\draw (493.05,296.05) node   [align=left] {\begin{minipage}[lt]{63.2pt}\setlength\topsep{0pt}
\begin{center}

\end{center}

\end{minipage}};
% Text Node
\draw  [fill={rgb, 255:red, 254; green, 243; blue, 191 }  ,fill opacity=1 ]  (561.5,253.05) -- (657.5,253.05) -- (657.5,339.05) -- (561.5,339.05) -- cycle  ;
\draw (609.5,296.05) node   [align=left] {\begin{minipage}[lt]{63.2pt}\setlength\topsep{0pt}
\begin{center}

\end{center}

\end{minipage}};
% Text Node
\draw  [fill={rgb, 255:red, 248; green, 231; blue, 28 }  ,fill opacity=1 ]  (173.81,358.8) -- (270.81,358.8) -- (270.81,444.8) -- (173.81,444.8) -- cycle  ;
\draw (222.31,401.8) node   [align=left] {\begin{minipage}[lt]{64.04pt}\setlength\topsep{0pt}
\begin{center}
\textbf{7 features}\\{\small \textit{Bounding Box,}}\\{\small \textit{Head Pose}}\\
\end{center}

\end{minipage}};
% Text Node
\draw (408.01,390.28) node [anchor=north west][inner sep=0.75pt]   [align=left] {\textbf{...}};
% Text Node
\draw  [fill={rgb, 255:red, 233; green, 166; blue, 174 }  ,fill opacity=1 ]  (289.05,358.8) -- (385.05,358.8) -- (385.05,444.8) -- (289.05,444.8) -- cycle  ;
\draw (337.05,401.8) node   [align=left] {\begin{minipage}[lt]{63.2pt}\setlength\topsep{0pt}
\begin{center}
No features Detected 
\end{center}

\end{minipage}};
% Text Node
\draw  [fill={rgb, 255:red, 254; green, 243; blue, 191 }  ,fill opacity=1 ]  (443.95,358.8) -- (539.95,358.8) -- (539.95,444.8) -- (443.95,444.8) -- cycle  ;
\draw (491.95,401.8) node   [align=left] {\begin{minipage}[lt]{63.2pt}\setlength\topsep{0pt}
\begin{center}

\end{center}

\end{minipage}};
% Text Node
\draw  [fill={rgb, 255:red, 254; green, 243; blue, 191 }  ,fill opacity=1 ]  (560.4,358.8) -- (656.4,358.8) -- (656.4,444.8) -- (560.4,444.8) -- cycle  ;
\draw (608.4,401.8) node   [align=left] {\begin{minipage}[lt]{63.2pt}\setlength\topsep{0pt}
\begin{center}

\end{center}

\end{minipage}};
% Text Node
\draw  [fill={rgb, 255:red, 203; green, 199; blue, 199 }  ,fill opacity=1 ]  (2.9,495.06) -- (138.9,495.06) -- (138.9,526.06) -- (2.9,526.06) -- cycle  ;
\draw (70.9,510.56) node   [align=left] {\begin{minipage}[lt]{90.66pt}\setlength\topsep{0pt}
\begin{center}
Output
\end{center}

\end{minipage}};
% Text Node
\draw  [fill={rgb, 255:red, 245; green, 239; blue, 166 }  ,fill opacity=1 ]  (152.58,494.9) -- (261.58,494.9) -- (261.58,529.9) -- (152.58,529.9) -- cycle  ;
\draw (207.08,512.4) node   [align=left] {\begin{minipage}[lt]{72pt}\setlength\topsep{0pt}
\begin{center}
$\displaystyle \left[ x_{1}^{1} ,\ ...,\ x_{d}^{1}\right]$\\
\end{center}

\end{minipage}};
% Text Node
\draw (387.91,510.75) node [anchor=north west][inner sep=0.75pt]   [align=left] {\textbf{...}};
% Text Node
\draw  [fill={rgb, 255:red, 233; green, 166; blue, 174 }  ,fill opacity=1 ]  (269.74,494.93) -- (378.74,494.93) -- (378.74,529.93) -- (269.74,529.93) -- cycle  ;
\draw (324.24,512.43) node   [align=left] {\begin{minipage}[lt]{72pt}\setlength\topsep{0pt}
\begin{center}
$\displaystyle [ NA,\ ...,NA]$\\
\end{center}

\end{minipage}};
% Text Node
\draw  [fill={rgb, 255:red, 254; green, 243; blue, 191 }  ,fill opacity=1 ]  (411.18,494.93) -- (520.18,494.93) -- (520.18,529.93) -- (411.18,529.93) -- cycle  ;
\draw (465.68,512.43) node   [align=left] {\begin{minipage}[lt]{72pt}\setlength\topsep{0pt}
\begin{center}
$\displaystyle \left[ x_{1}^{k-1} ,\ ...,\ x_{d}^{k-1}\right]$\\
\end{center}

\end{minipage}};
% Text Node
\draw  [fill={rgb, 255:red, 254; green, 243; blue, 191 }  ,fill opacity=1 ]  (527.92,494.93) -- (636.92,494.93) -- (636.92,529.93) -- (527.92,529.93) -- cycle  ;
\draw (582.42,512.43) node   [align=left] {\begin{minipage}[lt]{72pt}\setlength\topsep{0pt}
\begin{center}
$\displaystyle \left[ x_{1}^{k} ,\ ...,\ x_{d}^{k}\right]$\\
\end{center}

\end{minipage}};

\end{tikzpicture}

%% file: latex/figures/feature_engineering.tex
\tikzset{every picture/.style={line width=0.75pt}} %set default line width to 0.75pt        

\begin{tikzpicture}[x=0.75pt,y=0.75pt,yscale=-1,xscale=1]
%uncomment if require: \path (0,486); %set diagram left start at 0, and has height of 486

%Flowchart: Alternative Process [id:dp14193802287451818] 
\draw  [fill={rgb, 255:red, 208; green, 2; blue, 27 }  ,fill opacity=1 ] (123.4,74.84) .. controls (123.4,68.93) and (128.19,64.14) .. (134.1,64.14) -- (234.88,64.14) .. controls (240.79,64.14) and (245.58,68.93) .. (245.58,74.84) -- (245.58,114.6) .. controls (245.58,120.51) and (240.79,125.3) .. (234.88,125.3) -- (134.1,125.3) .. controls (128.19,125.3) and (123.4,120.51) .. (123.4,114.6) -- cycle ;
%Image [id:dp19974861338898897] 
\draw (567.42,25.81) node  {\includegraphics[width=34.23pt,height=31.12pt]{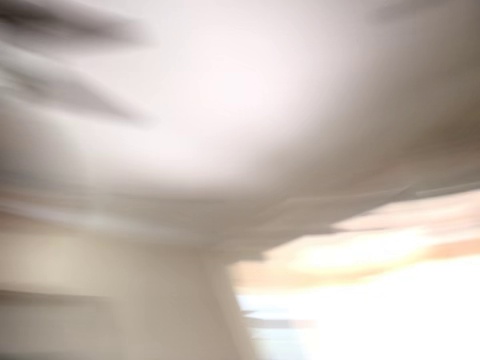}};
%Image [id:dp24640213526668542] 
\draw (274.29,26.13) node  {\includegraphics[width=39.36pt,height=32.13pt]{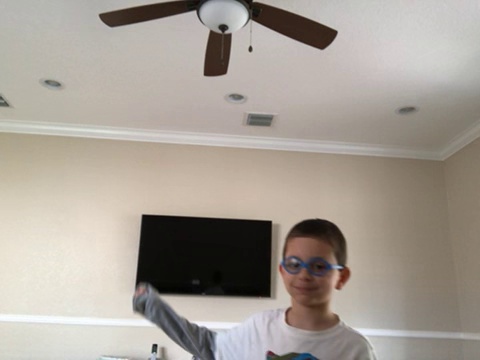}};
%Image [id:dp7026706635164364] 
\draw (514.56,25.47) node  {\includegraphics[width=34.52pt,height=31.45pt]{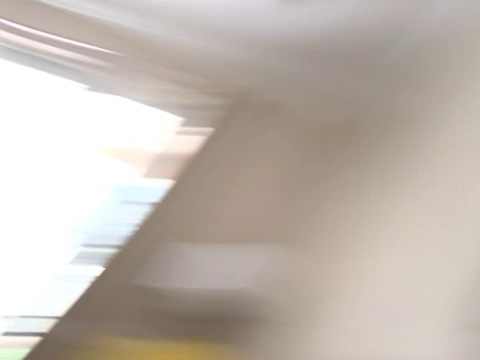}};
%Image [id:dp4844897915351045] 
\draw (672.61,26.27) node  {\includegraphics[width=33.33pt,height=31.86pt]{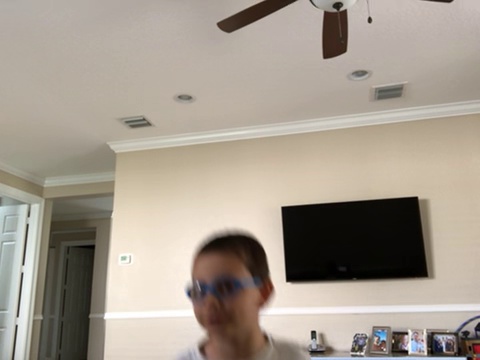}};
%Image [id:dp3645789114249305] 
\draw (619.85,26.46) node  {\includegraphics[width=33.67pt,height=31.72pt]{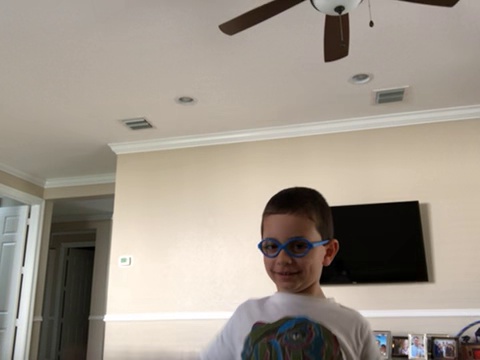}};
%Image [id:dp3700255751234671] 
\draw (459.13,25.79) node  {\includegraphics[width=38.12pt,height=31pt]{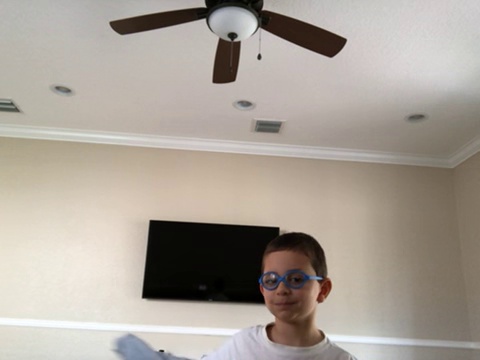}};
%Image [id:dp9531968794331207] 
\draw (397.55,26.22) node  {\includegraphics[width=39.79pt,height=32.27pt]{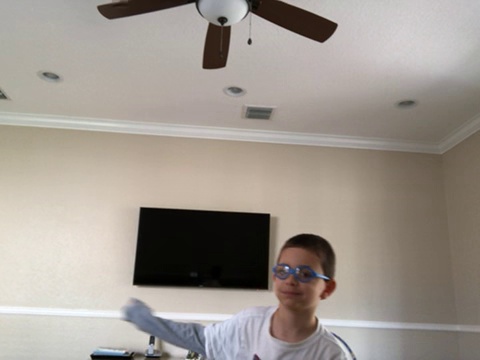}};
%Image [id:dp07984098736309186] 
\draw (336.19,26.14) node  {\includegraphics[width=39.03pt,height=32.2pt]{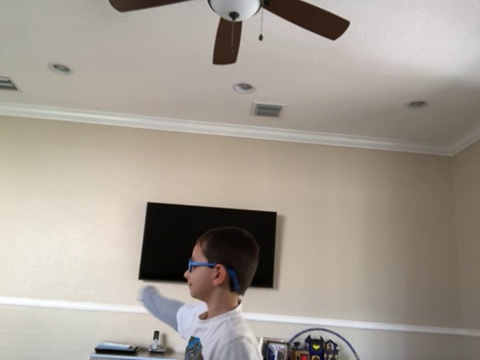}};
%Image [id:dp17861173199673308] 
\draw (152.42,26.32) node  {\includegraphics[width=39.47pt,height=32.42pt]{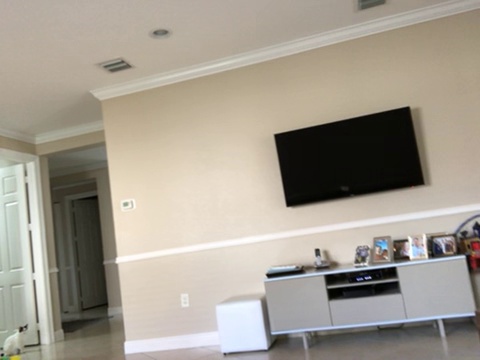}};
%Image [id:dp891653044866485] 
\draw (213.42,26.68) node  {\includegraphics[width=38.66pt,height=32.96pt]{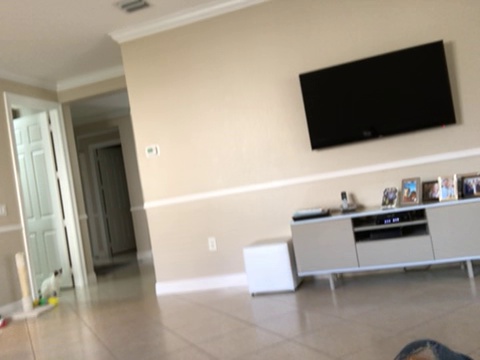}};
%Image [id:dp5842638446048649] 
\draw (567.99,92.86) node  {\includegraphics[width=34.23pt,height=31.12pt]{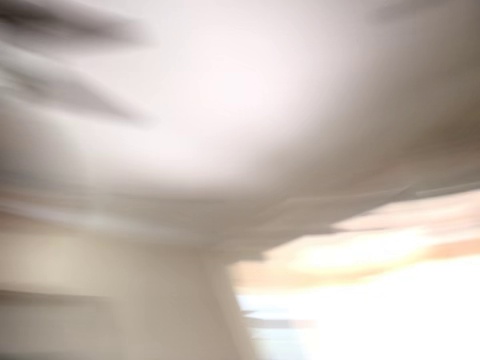}};
%Image [id:dp6775867774402147] 
\draw (274.86,93.18) node  {\includegraphics[width=39.36pt,height=32.13pt]{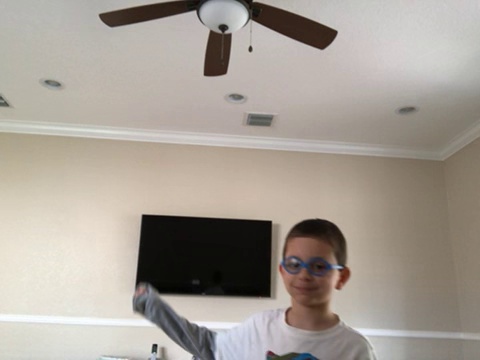}};
%Image [id:dp636748505708826] 
\draw (515.13,92.52) node  {\includegraphics[width=34.52pt,height=31.45pt]{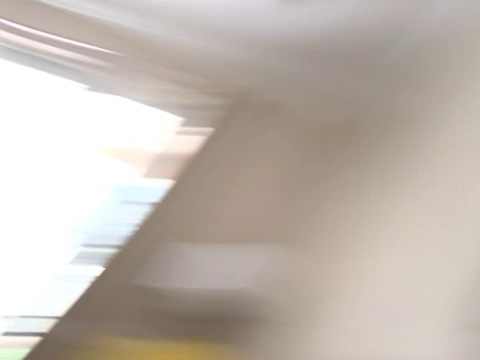}};
%Image [id:dp3014711455226766] 
\draw (673.18,93.32) node  {\includegraphics[width=33.33pt,height=31.86pt]{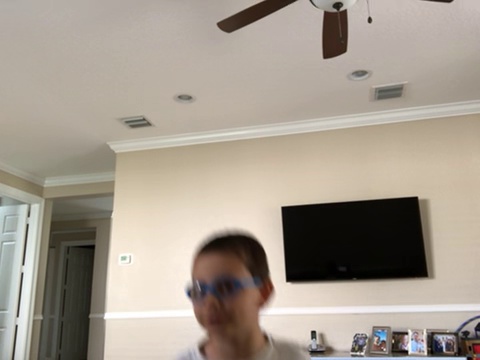}};
%Image [id:dp921848538249767] 
\draw (620.42,93.51) node  {\includegraphics[width=33.67pt,height=31.72pt]{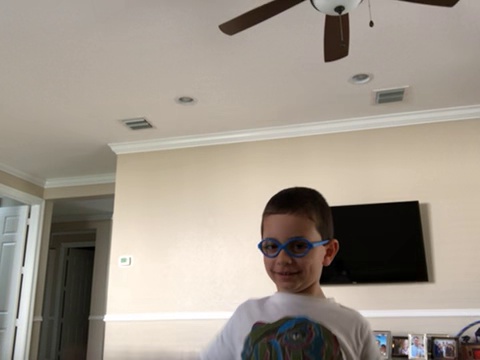}};
%Image [id:dp08138998692237975] 
\draw (458.94,92.35) node  {\includegraphics[width=39.25pt,height=30.28pt]{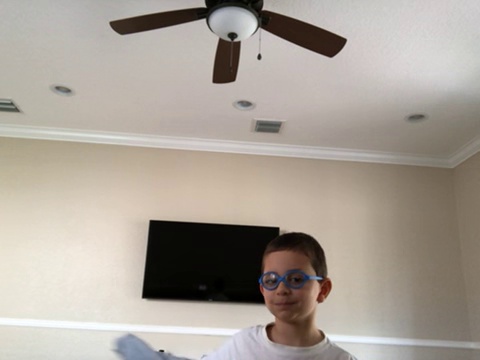}};
%Image [id:dp9824702761414907] 
\draw (395.53,92.77) node  {\includegraphics[width=40.98pt,height=31.52pt]{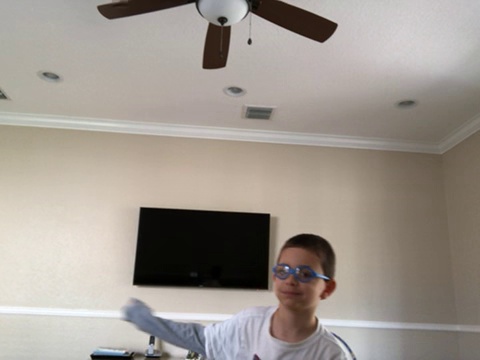}};
%Image [id:dp4067658816774009] 
\draw (336.76,93.19) node  {\includegraphics[width=39.03pt,height=32.2pt]{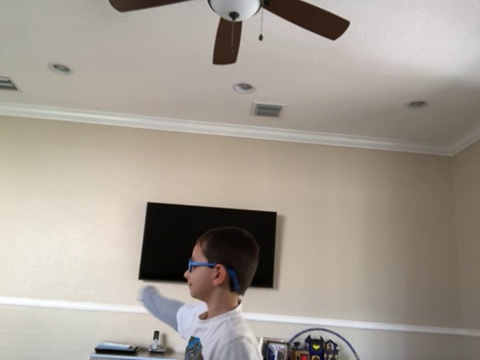}};
%Image [id:dp09924906994781568] 
\draw (152.99,93.37) node  {\includegraphics[width=39.47pt,height=32.42pt]{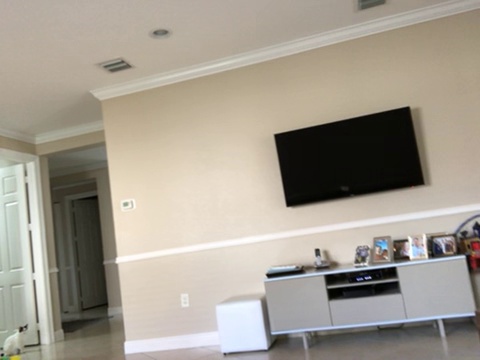}};
%Image [id:dp2118922804784673] 
\draw (213.99,93.73) node  {\includegraphics[width=38.66pt,height=32.96pt]{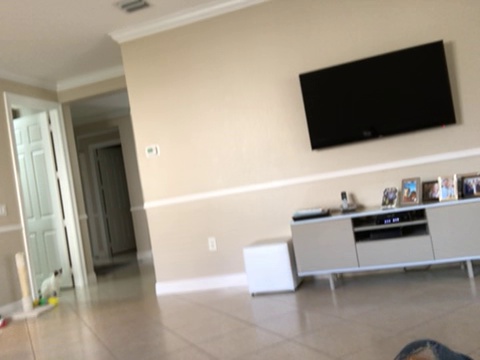}};
%Flowchart: Alternative Process [id:dp6471140528320058] 
\draw  [fill={rgb, 255:red, 208; green, 2; blue, 27 }  ,fill opacity=1 ] (422.29,141.52) .. controls (422.29,135.87) and (426.86,131.3) .. (432.51,131.3) -- (519.34,131.3) .. controls (524.99,131.3) and (529.56,135.87) .. (529.56,141.52) -- (529.56,179.47) .. controls (529.56,185.11) and (524.99,189.69) .. (519.34,189.69) -- (432.51,189.69) .. controls (426.86,189.69) and (422.29,185.11) .. (422.29,179.47) -- cycle ;
%Image [id:dp9831679350839253] 
\draw (502.03,161.27) node  {\includegraphics[width=34.47pt,height=29.71pt]{latex/figures/frame_85.jpg}};
%Image [id:dp8573839093515743] 
\draw (201.81,160.47) node  {\includegraphics[width=39.62pt,height=30.67pt]{latex/figures/frame_28.jpg}};
%Image [id:dp9622733400228485] 
\draw (448.81,160.95) node  {\includegraphics[width=34.75pt,height=30.03pt]{latex/figures/frame_14.jpg}};
%Image [id:dp02267624851812644] 
\draw (613.03,161.71) node  {\includegraphics[width=33.56pt,height=30.42pt]{latex/figures/frame_1.jpg}};
%Image [id:dp19668766938401694] 
\draw (559.91,161.89) node  {\includegraphics[width=33.9pt,height=30.28pt]{latex/figures/frame_2.jpg}};
%Image [id:dp6611851001352218] 
\draw (388.15,159.68) node  {\includegraphics[width=39.51pt,height=28.91pt]{latex/figures/frame_27.jpg}};
%Image [id:dp5848627158569499] 
\draw (326.36,160.08) node  {\includegraphics[width=41.25pt,height=30.09pt]{latex/figures/frame_26.jpg}};
%Image [id:dp5359662425761802] 
\draw (264.12,160.49) node  {\includegraphics[width=39.29pt,height=30.74pt]{latex/figures/frame_25.jpg}};
%Image [id:dp17786251114284823] 
\draw (191.58,245.07) node  {\includegraphics[width=40.77pt,height=33.56pt]{latex/figures/frame_28.jpg}};
%Image [id:dp8973410406448141] 
\draw (616.36,244.75) node  {\includegraphics[width=37.56pt,height=32.48pt]{latex/figures/frame_1.jpg}};
%Image [id:dp6534572475178761] 
\draw (556.9,244.94) node  {\includegraphics[width=37.94pt,height=32.33pt]{latex/figures/frame_2.jpg}};
%Image [id:dp18079059418362387] 
\draw (379.54,245.09) node  {\includegraphics[width=40.66pt,height=33.29pt]{latex/figures/frame_27.jpg}};
%Image [id:dp1618265660608844] 
\draw (316.85,244.64) node  {\includegraphics[width=42.45pt,height=32.92pt]{latex/figures/frame_26.jpg}};
%Image [id:dp1577171307382652] 
\draw (253.75,245.08) node  {\includegraphics[width=40.43pt,height=33.63pt]{latex/figures/frame_25.jpg}};
%Shape: Triangle [id:dp6989736534013491] 
\draw  [fill={rgb, 255:red, 126; green, 211; blue, 33 }  ,fill opacity=1 ] (300.4,210.7) -- (184,200.7) -- (414.4,200.7) -- cycle ;
%Shape: Triangle [id:dp8520518288990797] 
\draw  [fill={rgb, 255:red, 126; green, 211; blue, 33 }  ,fill opacity=1 ] (585.21,210.7) -- (536,200.7) -- (633.4,200.7) -- cycle ;
%Shape: Triangle [id:dp258793092651062] 
\draw  [fill={rgb, 255:red, 126; green, 211; blue, 33 }  ,fill opacity=1 ] (599.92,322.38) -- (527.84,312.02) -- (670.52,312.02) -- cycle ;
%Shape: Triangle [id:dp8962766188202751] 
\draw  [fill={rgb, 255:red, 126; green, 211; blue, 33 }  ,fill opacity=1 ] (218.6,323.39) -- (146.52,313.03) -- (289.2,313.03) -- cycle ;
%Shape: Triangle [id:dp5996706569480474] 
\draw  [fill={rgb, 255:red, 126; green, 211; blue, 33 }  ,fill opacity=1 ] (406.73,323.39) -- (334.64,313.03) -- (477.33,313.03) -- cycle ;

% Text Node
\draw  [fill={rgb, 255:red, 184; green, 233; blue, 134 }  ,fill opacity=1 ]  (130.07,282.76) -- (217.07,282.76) -- (217.07,305.76) -- (130.07,305.76) -- cycle  ;
\draw (133.07,287.16) node [anchor=north west][inner sep=0.75pt]  [font=\fontsize{0.62em}{0.74em}\selectfont]  {${\textstyle X^{1} =\left[ x_{1}^{1} ,\ ...,\ x_{d}^{1}\right]}$};
% Text Node
\draw  [fill={rgb, 255:red, 184; green, 233; blue, 134 }  ,fill opacity=1 ]  (221.42,282.76) -- (308.42,282.76) -- (308.42,305.76) -- (221.42,305.76) -- cycle  ;
\draw (224.42,287.16) node [anchor=north west][inner sep=0.75pt]  [font=\fontsize{0.62em}{0.74em}\selectfont]  {${\textstyle X^{2} =\left[ x_{1}^{2} ,\ ...,\ x_{d}^{2}\right]}$};
% Text Node
\draw  [fill={rgb, 255:red, 184; green, 233; blue, 134 }  ,fill opacity=1 ]  (316.01,282.78) -- (403.01,282.78) -- (403.01,305.78) -- (316.01,305.78) -- cycle  ;
\draw (319.01,287.18) node [anchor=north west][inner sep=0.75pt]  [font=\fontsize{0.62em}{0.74em}\selectfont]  {${\textstyle X^{3} =\left[ x_{1}^{3} ,\ ...,\ x_{d}^{3}\right]}$};
% Text Node
\draw  [fill={rgb, 255:red, 184; green, 233; blue, 134 }  ,fill opacity=1 ]  (407.68,282.79) -- (494.68,282.79) -- (494.68,305.79) -- (407.68,305.79) -- cycle  ;
\draw (410.68,287.19) node [anchor=north west][inner sep=0.75pt]  [font=\fontsize{0.62em}{0.74em}\selectfont]  {${\textstyle X^{4} =\left[ x_{1}^{4} ,\ ...,\ x_{d}^{4}\right]}$};
% Text Node
\draw  [fill={rgb, 255:red, 184; green, 233; blue, 134 }  ,fill opacity=1 ]  (509.37,281.78) -- (596.37,281.78) -- (596.37,304.78) -- (509.37,304.78) -- cycle  ;
\draw (512.37,286.18) node [anchor=north west][inner sep=0.75pt]  [font=\fontsize{0.62em}{0.74em}\selectfont]  {${\textstyle X^{5} =\left[ x_{1}^{5} ,\ ...,\ x_{d}^{5}\right]}$};
% Text Node
\draw  [fill={rgb, 255:red, 184; green, 233; blue, 134 }  ,fill opacity=1 ]  (602.39,281.78) -- (689.39,281.78) -- (689.39,304.78) -- (602.39,304.78) -- cycle  ;
\draw (605.39,286.18) node [anchor=north west][inner sep=0.75pt]  [font=\fontsize{0.62em}{0.74em}\selectfont]  {${\textstyle X^{6} =\left[ x_{1}^{6} ,\ ...,\ x_{d}^{6}\right]}$};
% Text Node
\draw  [fill={rgb, 255:red, 184; green, 233; blue, 134 }  ,fill opacity=1 ]  (132.1,330.57) -- (306.1,330.57) -- (306.1,356.57) -- (132.1,356.57) -- cycle  ;
\draw (135.1,334.97) node [anchor=north west][inner sep=0.75pt]  [font=\fontsize{0.62em}{0.74em}\selectfont]  {${\textstyle Z^{1} =\left[\left( x_{1}^{1} +x_{1}^{2}\right) /2,\ ...,\ \left( x_{d}^{1} +x_{d}^{2}\right) /2\right]}$};
% Text Node
\draw  [fill={rgb, 255:red, 184; green, 233; blue, 134 }  ,fill opacity=1 ]  (320.34,330.57) -- (494.34,330.57) -- (494.34,356.57) -- (320.34,356.57) -- cycle  ;
\draw (323.34,334.97) node [anchor=north west][inner sep=0.75pt]  [font=\fontsize{0.62em}{0.74em}\selectfont]  {${\textstyle Z^{2} =\left[\left( x_{1}^{3} +x_{1}^{4}\right) /2,\ ...,\ \left( x_{d}^{3} +x_{d}^{4}\right) /2\right]}$};
% Text Node
\draw  [fill={rgb, 255:red, 184; green, 233; blue, 134 }  ,fill opacity=1 ]  (512.92,330.43) -- (686.92,330.43) -- (686.92,356.43) -- (512.92,356.43) -- cycle  ;
\draw (515.92,334.83) node [anchor=north west][inner sep=0.75pt]  [font=\fontsize{0.62em}{0.74em}\selectfont]  {${\textstyle Z^{3} =\left[\left( x_{1}^{5} +x_{1}^{6}\right) /2,\ ...,\ \left( x_{d}^{5} +x_{d}^{6}\right) /2\right]}$};
% Text Node
\draw    (0.2,5.25) -- (120.2,5.25) -- (120.2,39.25) -- (0.2,39.25) -- cycle  ;
\draw (60.2,22.25) node   [align=left] {\begin{minipage}[lt]{79.15pt}\setlength\topsep{0pt}
\begin{center}
{\footnotesize Raw Input \ (10 fps)}
\end{center}

\end{minipage}};
% Text Node
\draw    (0.2,43.25) -- (120.2,43.25) -- (120.2,132.25) -- (0.2,132.25) -- cycle  ;
\draw (60.2,87.75) node   [align=left] {\begin{minipage}[lt]{79.15pt}\setlength\topsep{0pt}
\begin{center}
{\footnotesize Truncate unstable frames at the }\\{\footnotesize beginning and the end}
\end{center}

\end{minipage}};
% Text Node
\draw    (1.2,139.25) -- (121.2,139.25) -- (121.2,208.25) -- (1.2,208.25) -- cycle  ;
\draw (61.2,173.75) node   [align=left] {\begin{minipage}[lt]{79.15pt}\setlength\topsep{0pt}
\begin{center}
{\footnotesize Delete windows }\\{\footnotesize where the camera }\\{\footnotesize is moving}
\end{center}

\end{minipage}};
% Text Node
\draw    (1.2,214.25) -- (121.2,214.25) -- (121.2,285.25) -- (1.2,285.25) -- cycle  ;
\draw (61.2,249.75) node   [align=left] {\begin{minipage}[lt]{79.15pt}\setlength\topsep{0pt}
\begin{center}
{\footnotesize Concatenate }\\{\footnotesize informative windows of features}
\end{center}

\end{minipage}};
% Text Node
\draw    (1.2,292) -- (120.2,292) -- (120.2,346) -- (1.2,346) -- cycle  ;
\draw (60.7,319) node   [align=left] {\begin{minipage}[lt]{78.47pt}\setlength\topsep{0pt}
\begin{center}
{\footnotesize Downsampling by }\\{\footnotesize averaging}
\end{center}

\end{minipage}};
% Text Node
\draw    (0.2,351.9) -- (119.2,351.9) -- (119.2,399.9) -- (0.2,399.9) -- cycle  ;
\draw (59.7,375.9) node   [align=left] {\begin{minipage}[lt]{78.47pt}\setlength\topsep{0pt}
\begin{center}
{\footnotesize Preprocessed Input (5 fps)}
\end{center}

\end{minipage}};
% Text Node
\draw  [fill={rgb, 255:red, 184; green, 233; blue, 134 }  ,fill opacity=1 ]  (151.73,368.19) -- (239.73,368.19) -- (239.73,391.19) -- (151.73,391.19) -- cycle  ;
\draw (154.73,372.59) node [anchor=north west][inner sep=0.75pt]  [font=\fontsize{0.62em}{0.74em}\selectfont]  {${\textstyle Z^{1} =\left[ z_{1}^{1} \ ,\ ...,\ z_{d}^{1}\right]}$};
% Text Node
\draw  [fill={rgb, 255:red, 184; green, 233; blue, 134 }  ,fill opacity=1 ]  (260.18,369.19) -- (348.18,369.19) -- (348.18,392.19) -- (260.18,392.19) -- cycle  ;
\draw (263.18,373.59) node [anchor=north west][inner sep=0.75pt]  [font=\fontsize{0.62em}{0.74em}\selectfont]  {${\textstyle Z^{2} =\left[ z_{1}^{2} \ ,\ ...,\ z_{d}^{2}\right]}$};
% Text Node
\draw  [fill={rgb, 255:red, 184; green, 233; blue, 134 }  ,fill opacity=1 ]  (367.55,369.19) -- (455.55,369.19) -- (455.55,392.19) -- (367.55,392.19) -- cycle  ;
\draw (370.55,373.59) node [anchor=north west][inner sep=0.75pt]  [font=\fontsize{0.62em}{0.74em}\selectfont]  {${\textstyle Z^{3} =\left[ z_{1}^{3} \ ,\ ...,\ z_{d}^{3}\right]}$};

\end{tikzpicture}

%% file: normalization.tex
\tikzset{every picture/.style={line width=0.75pt}} %set default line width to 0.75pt        

\begin{tikzpicture}[x=0.75pt,y=0.75pt,yscale=-1,xscale=1]
%uncomment if require: \path (0,513); %set diagram left start at 0, and has height of 513

%Image [id:dp9774434845672941] 
\draw (328.5,215.77) node  {\includegraphics[width=495.59pt,height=326.25pt]{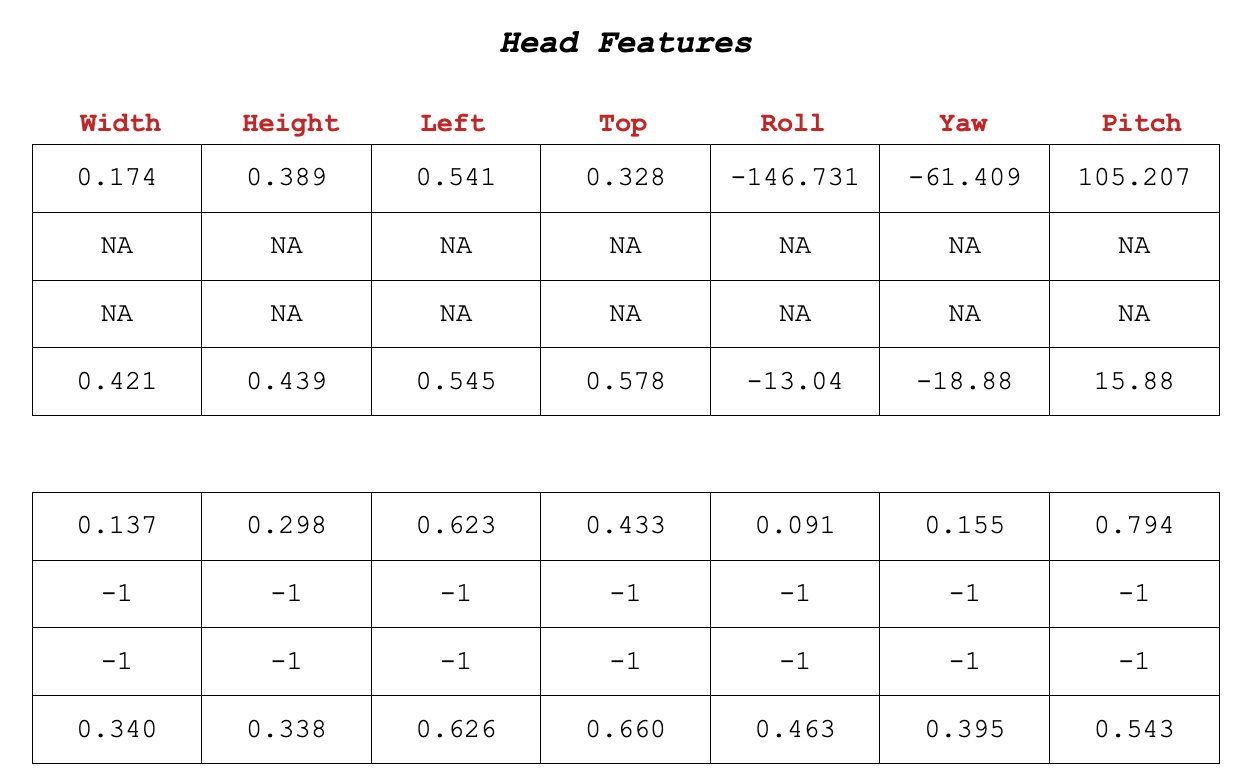}};
%Shape: Rectangle [id:dp6043638253100345] 
\draw  [color={rgb, 255:red, 74; green, 144; blue, 226 }  ,draw opacity=1 ][line width=2.25]  (15.89,116) -- (648.89,116) -- (648.89,193.05) -- (15.89,193.05) -- cycle ;
%Shape: Rectangle [id:dp8770490699096793] 
\draw  [color={rgb, 255:red, 74; green, 144; blue, 226 }  ,draw opacity=1 ][line width=2.25]  (15.89,310.05) -- (647.89,310.05) -- (647.89,385.05) -- (15.89,385.05) -- cycle ;
%Straight Lines [id:da9489144251331829] 
\draw [color={rgb, 255:red, 65; green, 117; blue, 5 }  ,draw opacity=1 ][line width=1.5]    (15.89,46.05) -- (102.89,46.05) ;
%Straight Lines [id:da882870634268138] 
\draw [color={rgb, 255:red, 65; green, 117; blue, 5 }  ,draw opacity=1 ][line width=1.5]    (274.89,45.05) -- (369.89,45.05) ;
%Straight Lines [id:da8987606928656555] 
\draw [color={rgb, 255:red, 65; green, 117; blue, 5 }  ,draw opacity=1 ][line width=1.5]    (15.89,46.05) -- (15.89,68.05) ;
%Straight Lines [id:da4809556927722838] 
\draw [color={rgb, 255:red, 65; green, 117; blue, 5 }  ,draw opacity=1 ][line width=1.5]    (369.89,45.05) -- (369.89,67.05) ;
%Straight Lines [id:da9096577539786723] 
\draw [color={rgb, 255:red, 65; green, 117; blue, 5 }  ,draw opacity=1 ][line width=1.5]    (381.89,46.05) -- (381.89,68.05) ;
%Straight Lines [id:da24062391895858992] 
\draw [color={rgb, 255:red, 65; green, 117; blue, 5 }  ,draw opacity=1 ][line width=1.5]    (650.89,45.05) -- (650.89,67.05) ;
%Straight Lines [id:da038231037433362314] 
\draw [color={rgb, 255:red, 65; green, 117; blue, 5 }  ,draw opacity=1 ][line width=1.5]    (381.89,46.05) -- (408.89,46.05) ;
%Straight Lines [id:da3731185255515528] 
\draw [color={rgb, 255:red, 65; green, 117; blue, 5 }  ,draw opacity=1 ][line width=1.5]    (623.89,45.05) -- (650.89,45.05) ;
%Flowchart: Merge [id:dp022637191303015003] 
\draw  [fill={rgb, 255:red, 155; green, 155; blue, 155 }  ,fill opacity=1 ] (17.89,239) -- (646.89,239) -- (332.39,260.38) -- cycle ;

% Text Node
\draw (109,37) node [anchor=north west][inner sep=0.75pt]   [align=left] {{\fontfamily{pcr}\selectfont \textcolor[rgb]{0.25,0.46,0.02}{\textit{\textbf{ranges from 0 to 1}}}}};
% Text Node
\draw (413,37) node [anchor=north west][inner sep=0.75pt]   [align=left] {{\fontfamily{pcr}\selectfont \textit{\textbf{\textcolor[rgb]{0.25,0.46,0.02}{ranges from -180 to 180}}}}};

\end{tikzpicture}

%% file: fusion.tex
\tikzset{every picture/.style={line width=0.75pt}} %set default line width to 0.75pt        

\begin{tikzpicture}[x=0.75pt,y=0.75pt,yscale=-1,xscale=1]
%uncomment if require: \path (0,584); %set diagram left start at 0, and has height of 584

%Bend Up Arrow [id:dp5573481551848436] 
\draw  [fill={rgb, 255:red, 184; green, 233; blue, 134 }  ,fill opacity=1 ] (300.57,139.48) -- (455.78,139.48) -- (455.78,130.18) -- (451.13,130.18) -- (460.43,117.78) -- (469.73,130.18) -- (465.08,130.18) -- (465.08,148.78) -- (300.57,148.78) -- cycle ;
%Bend Up Arrow [id:dp33236104640148323] 
\draw  [color={rgb, 255:red, 0; green, 0; blue, 0 }  ,draw opacity=1 ][fill={rgb, 255:red, 184; green, 233; blue, 134 }  ,fill opacity=1 ] (300.45,39.48) -- (456.49,39.48) -- (456.49,48.18) -- (452.14,48.18) -- (460.84,59.78) -- (469.54,48.18) -- (465.19,48.18) -- (465.19,30.78) -- (300.45,30.78) -- cycle ;
%Right Arrow [id:dp7092644536672368] 
\draw  [fill={rgb, 255:red, 184; green, 233; blue, 134 }  ,fill opacity=1 ] (300.45,81.72) -- (420.3,81.72) -- (420.3,76.78) -- (439.88,86.67) -- (420.3,96.56) -- (420.3,91.61) -- (300.45,91.61) -- cycle ;
%Right Arrow [id:dp6469191605367872] 
\draw  [fill={rgb, 255:red, 184; green, 233; blue, 134 }  ,fill opacity=1 ] (299.95,256.29) -- (419.94,256.29) -- (419.94,251.16) -- (439.54,261.42) -- (419.94,271.67) -- (419.94,266.54) -- (299.95,266.54) -- cycle ;
%Right Arrow [id:dp34467362638077526] 
\draw  [fill={rgb, 255:red, 184; green, 233; blue, 134 }  ,fill opacity=1 ] (300.85,203.91) -- (422.73,203.91) -- (422.73,198.78) -- (442.64,209.03) -- (422.73,219.28) -- (422.73,214.16) -- (300.85,214.16) -- cycle ;
%Right Arrow [id:dp8044476229111577] 
\draw  [fill={rgb, 255:red, 184; green, 233; blue, 134 }  ,fill opacity=1 ] (300.85,306.4) -- (422.73,306.4) -- (422.73,301.28) -- (442.64,311.53) -- (422.73,321.78) -- (422.73,316.65) -- (300.85,316.65) -- cycle ;
%Right Arrow [id:dp9797618284662994] 
\draw  [fill={rgb, 255:red, 184; green, 233; blue, 134 }  ,fill opacity=1 ] (500.15,260.29) -- (548.84,260.29) -- (548.84,255.16) -- (575.12,265.42) -- (548.84,275.67) -- (548.84,270.54) -- (500.15,270.54) -- cycle ;
%Straight Lines [id:da12863414192269196] 
\draw [line width=1.5]    (131.09,13.34) -- (131.45,353.56) ;
%Right Arrow [id:dp8037210703107915] 
\draw  [fill={rgb, 255:red, 184; green, 233; blue, 134 }  ,fill opacity=1 ] (499.44,460.72) -- (548.12,460.72) -- (548.12,455.59) -- (574.41,465.84) -- (548.12,476.09) -- (548.12,470.97) -- (499.44,470.97) -- cycle ;
%Right Arrow [id:dp39876049375085776] 
\draw  [fill={rgb, 255:red, 184; green, 233; blue, 134 }  ,fill opacity=1 ] (306.04,460.72) -- (426.04,460.72) -- (426.04,455.59) -- (445.64,465.84) -- (426.04,476.09) -- (426.04,470.97) -- (306.04,470.97) -- cycle ;
%Right Arrow [id:dp4167128510169149] 
\draw  [fill={rgb, 255:red, 184; green, 233; blue, 134 }  ,fill opacity=1 ] (306.95,405.33) -- (428.83,405.33) -- (428.83,400.21) -- (448.73,410.46) -- (428.83,420.71) -- (428.83,415.58) -- (306.95,415.58) -- cycle ;
%Right Arrow [id:dp6477951600345084] 
\draw  [fill={rgb, 255:red, 184; green, 233; blue, 134 }  ,fill opacity=1 ] (304.95,520.83) -- (426.83,520.83) -- (426.83,515.7) -- (446.73,525.95) -- (426.83,536.21) -- (426.83,531.08) -- (304.95,531.08) -- cycle ;
%Straight Lines [id:da7643014252270415] 
\draw [line width=1.5]    (132.19,384.98) -- (133.28,564.32) ;

% Text Node
\draw  [fill={rgb, 255:red, 203; green, 199; blue, 199 }  ,fill opacity=1 ]  (139.09,22.88) -- (247.09,22.88) -- (247.09,53.88) -- (139.09,53.88) -- cycle  ;
\draw (193.09,38.38) node   [align=left] {\begin{minipage}[lt]{71.59pt}\setlength\topsep{0pt}
\begin{center}
Eye Model
\end{center}

\end{minipage}};
% Text Node
\draw  [fill={rgb, 255:red, 255; green, 255; blue, 255 }  ,fill opacity=1 ]  (20.21,190.76) -- (118.21,190.76) -- (118.21,241.76) -- (20.21,241.76) -- cycle  ;
\draw (69.21,216.26) node   [align=left] {\begin{minipage}[lt]{64.8pt}\setlength\topsep{0pt}
\begin{center}
Late \\Fusion
\end{center}

\end{minipage}};
% Text Node
\draw  [fill={rgb, 255:red, 203; green, 199; blue, 199 }  ,fill opacity=1 ]  (139.25,74.88) -- (248.25,74.88) -- (248.25,107.88) -- (139.25,107.88) -- cycle  ;
\draw (193.75,91.38) node   [align=left] {\begin{minipage}[lt]{71.67pt}\setlength\topsep{0pt}
\begin{center}
Face Model
\end{center}

\end{minipage}};
% Text Node
\draw  [fill={rgb, 255:red, 203; green, 199; blue, 199 }  ,fill opacity=1 ]  (141.09,125.88) -- (248.09,125.88) -- (248.09,158.88) -- (141.09,158.88) -- cycle  ;
\draw (194.59,142.38) node   [align=left] {\begin{minipage}[lt]{70.91pt}\setlength\topsep{0pt}
\begin{center}
Head Model
\end{center}

\end{minipage}};
% Text Node
\draw (259.8,28) node [anchor=north west][inner sep=0.75pt]   [align=left] {$\displaystyle O_{1}$};
% Text Node
\draw (258.8,80) node [anchor=north west][inner sep=0.75pt]   [align=left] {$\displaystyle O_{2}$};
% Text Node
\draw (260.9,135) node [anchor=north west][inner sep=0.75pt]   [align=left] {$\displaystyle O_{3}$};
% Text Node
\draw  [fill={rgb, 255:red, 179; green, 211; blue, 248 }  ,fill opacity=1 ]  (443.11,63) -- (629.11,63) -- (629.11,111) -- (443.11,111) -- cycle  ;
\draw (446.11,67) node [anchor=north west][inner sep=0.75pt]   [align=left] {$\displaystyle O_{final} =\ \frac{( O_{1} +O_{2} +O_{3})}{3}$};
% Text Node
\draw  [fill={rgb, 255:red, 203; green, 199; blue, 199 }  ,fill opacity=1 ]  (452.44,187.34) -- (486.35,187.34) -- (486.35,344.48) -- (452.44,344.48) -- cycle  ;
\draw (469.39,265.91) node  [rotate=-269.66,xslant=0] [align=left] {\begin{minipage}[lt]{104.73pt}\setlength\topsep{0pt}
\begin{center}
Linear Layer
\end{center}

\end{minipage}};
% Text Node
\draw  [fill={rgb, 255:red, 179; green, 211; blue, 248 }  ,fill opacity=1 ]  (591.11,247) -- (638.11,247) -- (638.11,273) -- (591.11,273) -- cycle  ;
\draw (594.11,251) node [anchor=north west][inner sep=0.75pt]   [align=left] {$\displaystyle O_{final}$};
% Text Node
\draw  [fill={rgb, 255:red, 248; green, 231; blue, 28 }  ,fill opacity=1 ]  (274.64, 15.11) circle [x radius= 30.41, y radius= 14.14]   ;
\draw (254.14,7.11) node [anchor=north west][inner sep=0.75pt]   [align=left] {\begin{minipage}[lt]{29.94pt}\setlength\topsep{0pt}
\begin{center}
\textit{Logits}
\end{center}

\end{minipage}};
% Text Node
\draw  [fill={rgb, 255:red, 255; green, 255; blue, 255 }  ,fill opacity=1 ]  (6.95,444.08) -- (117.95,444.08) -- (117.95,497.08) -- (6.95,497.08) -- cycle  ;
\draw (62.45,470.58) node   [align=left] {\begin{minipage}[lt]{73.49pt}\setlength\topsep{0pt}
\begin{center}
Intermediate\\Fusion
\end{center}

\end{minipage}};
% Text Node
\draw  [fill={rgb, 255:red, 179; green, 211; blue, 248 }  ,fill opacity=1 ]  (581.4,452.43) -- (628.4,452.43) -- (628.4,478.43) -- (581.4,478.43) -- cycle  ;
\draw (584.4,456.43) node [anchor=north west][inner sep=0.75pt]   [align=left] {$\displaystyle O_{final}$};
% Text Node
\draw (260.9,404.43) node [anchor=north west][inner sep=0.75pt]   [align=left] {$\displaystyle H_{1}$};
% Text Node
\draw (260.9,458.43) node [anchor=north west][inner sep=0.75pt]   [align=left] {$\displaystyle H_{2}$};
% Text Node
\draw (260.99,511.43) node [anchor=north west][inner sep=0.75pt]   [align=left] {$\displaystyle H_{3}$};
% Text Node
\draw  [fill={rgb, 255:red, 203; green, 199; blue, 199 }  ,fill opacity=1 ]  (453.96,390.77) -- (487.87,390.77) -- (487.87,547.91) -- (453.96,547.91) -- cycle  ;
\draw (470.92,469.34) node  [rotate=-269.66,xslant=0] [align=left] {\begin{minipage}[lt]{104.73pt}\setlength\topsep{0pt}
\begin{center}
MLP
\end{center}

\end{minipage}};
% Text Node
\draw  [fill={rgb, 255:red, 248; green, 231; blue, 28 }  ,fill opacity=1 ]  (276.83, 373.93) circle [x radius= 61.65, y radius= 29.27]   ;
\draw (234.41,355.23) node [anchor=north west][inner sep=0.75pt]  [rotate=-0.26] [align=left] {\begin{minipage}[lt]{59.98pt}\setlength\topsep{0pt}
\begin{center}
\textit{Final Hidden}\\\textit{Dimensions }
\end{center}

\end{minipage}};
% Text Node
\draw  [fill={rgb, 255:red, 203; green, 199; blue, 199 }  ,fill opacity=1 ]  (139.09,196.76) -- (247.09,196.76) -- (247.09,227.76) -- (139.09,227.76) -- cycle  ;
\draw (193.09,212.26) node   [align=left] {\begin{minipage}[lt]{71.59pt}\setlength\topsep{0pt}
\begin{center}
Eye Model
\end{center}

\end{minipage}};
% Text Node
\draw  [fill={rgb, 255:red, 203; green, 199; blue, 199 }  ,fill opacity=1 ]  (139.25,248.76) -- (248.25,248.76) -- (248.25,281.76) -- (139.25,281.76) -- cycle  ;
\draw (193.75,265.26) node   [align=left] {\begin{minipage}[lt]{71.67pt}\setlength\topsep{0pt}
\begin{center}
Face Model
\end{center}

\end{minipage}};
% Text Node
\draw  [fill={rgb, 255:red, 203; green, 199; blue, 199 }  ,fill opacity=1 ]  (141.09,299.76) -- (248.09,299.76) -- (248.09,332.76) -- (141.09,332.76) -- cycle  ;
\draw (194.59,316.26) node   [align=left] {\begin{minipage}[lt]{70.91pt}\setlength\topsep{0pt}
\begin{center}
Head Model
\end{center}

\end{minipage}};
% Text Node
\draw (259.8,201.89) node [anchor=north west][inner sep=0.75pt]   [align=left] {$\displaystyle O_{1}$};
% Text Node
\draw (258.8,253.89) node [anchor=north west][inner sep=0.75pt]   [align=left] {$\displaystyle O_{2}$};
% Text Node
\draw (260.9,308.89) node [anchor=north west][inner sep=0.75pt]   [align=left] {$\displaystyle O_{3}$};
% Text Node
\draw  [fill={rgb, 255:red, 248; green, 231; blue, 28 }  ,fill opacity=1 ]  (274.64, 189) circle [x radius= 30.41, y radius= 14.14]   ;
\draw (254.14,181) node [anchor=north west][inner sep=0.75pt]   [align=left] {\begin{minipage}[lt]{29.94pt}\setlength\topsep{0pt}
\begin{center}
\textit{Logits}
\end{center}

\end{minipage}};
% Text Node
\draw  [fill={rgb, 255:red, 203; green, 199; blue, 199 }  ,fill opacity=1 ]  (142.09,400.76) -- (250.09,400.76) -- (250.09,431.76) -- (142.09,431.76) -- cycle  ;
\draw (196.09,416.26) node   [align=left] {\begin{minipage}[lt]{71.59pt}\setlength\topsep{0pt}
\begin{center}
Eye Model
\end{center}

\end{minipage}};
% Text Node
\draw  [fill={rgb, 255:red, 203; green, 199; blue, 199 }  ,fill opacity=1 ]  (142.25,452.76) -- (251.25,452.76) -- (251.25,485.76) -- (142.25,485.76) -- cycle  ;
\draw (196.75,469.26) node   [align=left] {\begin{minipage}[lt]{71.67pt}\setlength\topsep{0pt}
\begin{center}
Face Model
\end{center}

\end{minipage}};
% Text Node
\draw  [fill={rgb, 255:red, 203; green, 199; blue, 199 }  ,fill opacity=1 ]  (144.09,503.76) -- (251.09,503.76) -- (251.09,536.76) -- (144.09,536.76) -- cycle  ;
\draw (197.59,520.26) node   [align=left] {\begin{minipage}[lt]{70.91pt}\setlength\topsep{0pt}
\begin{center}
Head Model
\end{center}

\end{minipage}};

\end{tikzpicture}